\documentclass[runningheads]{llncs}

\usepackage{eccv}

\usepackage{multirow}

\usepackage{eccvabbrv}

\usepackage{graphicx}
\usepackage{booktabs}

\usepackage[accsupp]{axessibility}  

\usepackage{hyperref}

\usepackage{orcidlink}
\usepackage[misc]{ifsym}

\begin{document}

\title{Training-free Cross-domain Few-shot Segmentation via Robust Semantic Representation and Matching} 

\titlerunning{Training-free CD-FSS via Robust Semantic Representation and Matching}

\author{Sujun Sun\inst{1,2}\orcidlink{0009-0002-0647-1148} \and
Mingwu Ren\inst{1,2}\orcidlink{0000-0001-5576-3281} \and
Haofeng Zhang\inst{1,2}\textsuperscript{(\Letter)}\orcidlink{0000-0002-4039-7618}}

\authorrunning{S. Sun et al.}

\institute{School of Computer Science and Engineering, Nanjing University of Science and Technology, China \and
State Key Laboratory of Intelligent Manufacturing of Advanced Construction Machinery, China\\
\email{\{egg, renmingwu, zhanghf\}@njust.edu.cn}}

\maketitle

\begin{abstract}
Cross-domain Few-shot Segmentation (CD-FSS) aims to tra-nsfer knowledge learned from source domain to distinct target domains, segmenting unseen target classes with only a few annotated samples. Although existing methods have made significant progress, they still rely on training or fine-tuning processes, which incur high computational costs and risk overfitting. We observe that when powerful and general-purpose vision foundation models are incorporated into these methods, their performance shows only marginal improvement or even degrades due to overfitting. To address this, we eliminate trainable parameters and propose a training-free framework to avoid both training overhead and overfitting. Built upon the self-supervised vision encoder DINOv3, our framework addresses cross-domain challenges through three core modules. First, the Semantic-aware Feature Re-fusion (SAFR) module identifies and re-fuses features that emphasize semantic patterns, generating representations with enhanced semantic discriminability. Additionally, the Adaptive Support Enhancement (ASE) module narrows semantic gaps between support and query through robust query information aggregation. Finally, the Hybrid Prototype Matching (HPM) module integrates matching results from diverse prototypes to adapt to varying semantic complexity across domains. Extensive experiments on four target domain datasets demonstrate that our method achieves state-of-the-art performance in CD-FSS without any training. Our code is available at \url{https://github.com/Sparkling-Water/RSRM}.
\end{abstract}

\begin{figure}[t]
    \centering
    \includegraphics[width=0.98\linewidth]{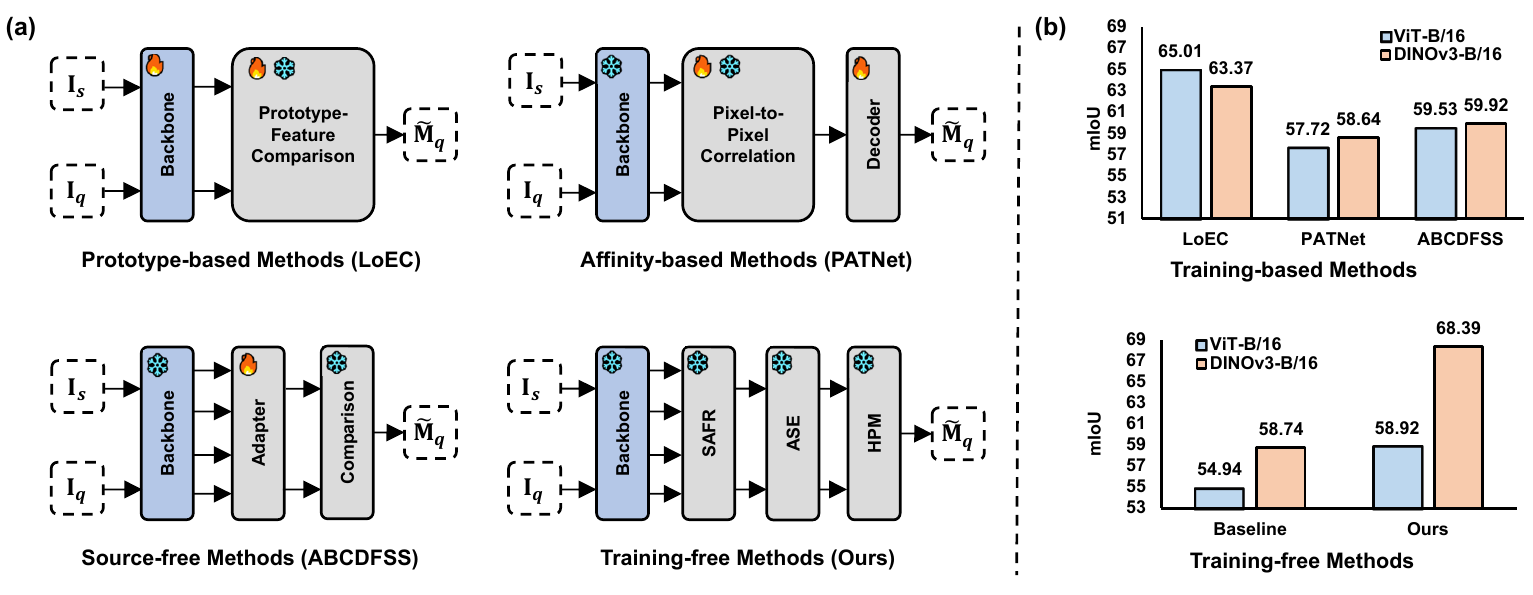}
    \vspace{-1ex}
    \caption{(a) Existing CD-FSS methods across three paradigms all involve different trainable parameters, while our method requires no additional training. (b) When DINOv3 is incorporated into three existing CD-FSS paradigms, their performance shows only marginal improvement or even degrades. In contrast, our training-free method eliminates overfitting risk and achieves state-of-the-art performance.}
    \vspace{-1ex}
    \label{fig:motivation}
\end{figure}

\section{Introduction}
\label{sec:intro}

As a fundamental task in computer vision, semantic segmentation has witnessed significant progress in recent years \cite{cheng2022masked,yu2024embedding,fu2025segman}. However, most existing methods remain limited to training classes with abundant pixel-level annotations and struggle to extend to novel classes. To address this bottleneck, Few-shot Segmentation (FSS) has been introduced \cite{sun2024vrp,peng2023hierarchical,fan2022self,wang2024rethinking}. It leverages meta-learning on base classes with abundant annotated samples to learn generalizable class-agnostic knowledge, enabling the segmentation of novel classes using only a few annotated samples. While achieving remarkable success on in-domain benchmarks, existing FSS methods suffer severe performance degradation when domain gaps exist between training and testing data, limiting their practical applicability.

To bridge domain gaps, Cross-domain Few-shot Segmentation (CD-FSS) \cite{lei2022cross} has recently attracted increasing attention. Current CD-FSS methods can be broadly categorized into three paradigms: prototype-based methods \cite{su2024domain,nie2024cross,liu2025devil,tong2025self}, affinity-based methods \cite{lei2022cross,tong2024lightweight,tong2025adapter}, and source-free fine-tuning methods \cite{herzog2024adapt}. Although these methods design various strategies to tackle cross-domain challenges, they still rely on source domain training or target domain fine-tuning, which incur high computational costs and risk overfitting. Furthermore, their backbone pretrained on ImageNet \cite{russakovsky2015imagenet} limits their performance. Meanwhile, vision foundation models (VFMs) \cite{caron2021emerging,oquab2023dinov2,kirillov2023segment}, pretrained on large-scale datasets, can provide powerful and universal visual representations and have demonstrated superior performance across various downstream tasks \cite{jose2025dinov2,yang2024depth,zhang2024bridge}. Recently, the self-supervised vision encoder DINOv3 \cite{simeoni2025dinov3} has further advanced numerous dense vision tasks \cite{yuan2025ad,li2025meddinov3,yang2025segdino}. A natural expectation is that leveraging these VFMs as feature extractors could substantially enhance the performance of CD-FSS.

To validate this, we design a simple baseline that directly uses the pretrained model to extract features and predicts masks via similarity between support prototypes and query features. As shown in \cref{fig:motivation}(b), the baseline results reveal the excellent cross-domain potential of VFMs. However, integrating the same model into three CD-FSS paradigms yields only marginal improvement or even degradation. This phenomenon highlights the key limitation of existing paradigms: the trainable parameters in these methods are prone to overfitting on the training data, and the strong initialization may further accelerate this process.

Based on these observations, we eliminate trainable parameters and adopt a training-free framework to avoid both training overhead and overfitting. Several FSS methods \cite{zhang2024bridge, liu2023matcher, zhang2023personalize} perform training-free adaptation on SAM \cite{kirillov2023segment} but suffer from the following limitations: 1) Due to the limited capability of SAM's encoder for cross-image semantic matching, these methods typically require additional VFMs. 2) They are designed for single-domain settings and perform poorly in cross-domain scenarios. Moreover, encoder–decoder methods \cite{sun2024vrp, zhang2023personalize} introduce more overfitting-prone parameters, generally resulting in inferior performance and efficiency compared with encoder-only methods \cite{fan2022self, liu2025devil, zhang2022feature} in CD-FSS. Therefore, we perform training-free adaptation on a single VFM encoder.

Specifically, we employ DINOv3 as the encoder and generalize it to cross-domain scenarios through three key modules. The core idea of our method is to generate discriminative semantic representations and perform robust matching across different domains. First, we observe that certain intermediate features in DINOv3 overemphasize local consistency and propagate to the final output via residual connections, limiting its semantic discriminability. To address this, we design a Semantic-aware Feature Re-fusion (SAFR) module, which adaptively identifies and re-fuses features that emphasize semantic patterns, enhancing the semantic representation quality. In addition, support and query images exhibit varying intra-class instance differences across domains, resulting in significant feature discrepancies. To improve semantic alignment, we design an Adaptive Support Enhancement (ASE) module. ASE adaptively identifies robust foreground and background regions in the query and aggregates robust query information into support features. Finally, we introduce a Hybrid Prototype Matching (HPM) module to handle the variations in intra-image semantic complexity across domains, generating final predictions by fusing the matching results of global, regional, and pixel-level prototypes. Our contributions are as follows:
\begin{itemize}
\item We identify that existing CD-FSS paradigms are prone to overfitting when combined with vision foundation models and propose to avoid both training overhead and overfitting in a training-free manner.
\item We propose a training-free framework that incorporates three novel modules: SAFR enhances semantic discriminability of features, ASE improves semantic consistency between support and query, and HPM adapts to varying semantic complexity across domains.
\item Extensive experiments on four standard target domain datasets demonstrate that our method significantly outperforms current state-of-the-art CD-FSS methods, even without any training.
\end{itemize}

\section{Related Works}
\label{sec:related}

\subsection{Few-shot Segmentation}
The goal of FSS is to segment novel classes in query images using only a few annotated examples. Based on how support samples are utilized, existing FSS methods can be broadly categorized into prototype-based and affinity-based methods. Prototype-based methods adopt global \cite{fan2022self,zhang2019canet} or local \cite{yang2020prototype,li2021adaptive} prototypes to represent target classes in support images and then match these prototypes with query features through cosine similarity or convolution operations to generate segmentation masks. In contrast, affinity-based methods \cite{peng2023hierarchical,xu2023self,xu2024eliminating} argue that prototype computation ignores spatial details, thus focusing on constructing pixel-level correlations between support and query features and decoding them to predict query masks. Recently, some works \cite{sun2024vrp,chang2024high,xu2025unlocking} have leveraged the well-learned knowledge of vision foundation models (\eg, DINOv2 \cite{oquab2023dinov2}, SAM \cite{kirillov2023segment} and SAM2 \cite{ravi2024sam}) to simplify the learning of FSS, achieving significant performance improvements. However, these methods assume that base and novel classes originate from the same domain and fail to generalize well to unseen domains.

\subsection{Cross-domain Few-shot Segmentation}
Compared with FSS, CD-FSS is a more realistic setting, requiring the model trained on the source domain to generalize to novel classes in different target domains. Similar to the categorization in FSS, existing CD-FSS methods can be mainly divided into three paradigms: prototype-based methods, affinity-based methods, and source-free fine-tuning methods. Prototype-based methods \cite{su2024domain,nie2024cross,liu2025devil,tong2025self,sun2026bridging} typically train the backbone on the source domain, introducing strategies such as domain perturbation or feature decomposition to prevent overfitting, and generate query masks via a prototype-feature matching module. Affinity-based methods \cite{lei2022cross,tong2024lightweight,tong2025adapter} freeze the pretrained backbone, transform original features into domain-agnostic representations, compute dense affinities between support and query features, and then generate query masks using a decoder trained on the source domain. Source-free fine-tuning methods \cite{herzog2024adapt} also freeze the pretrained backbone but introduce a set of learnable adaptation layers for its hierarchical features, which are then fine-tuned with limited target domain samples. All these methods rely on training or fine-tuning processes, which incur substantial computational costs and risk overfitting. In this paper, we discard both training and fine-tuning processes, and avoid training overhead and overfitting in a training-free manner.

\subsection{Training-free Few-shot Segmentation}
Some FSS methods \cite{liu2023matcher,zhang2023personalize,zhang2024bridge} and Few-shot Medical Image Segmentation (FSMIS) approaches \cite{zhu2025maup,liu2025synpo} have achieved promising performance by improving upon the class-agnostic segmentation framework of SAM \cite{kirillov2023segment} without introducing learnable parameters, thereby avoiding high training costs and the risk of overfitting. However, these methods typically require integration with additional vision foundation models, resulting in a more complex pipeline. Moreover, they are primarily designed for single-domain data (such as natural images or medical images), and their effectiveness in cross-domain scenarios remains unverified. In contrast, we aim to generalize a single pretrained vision encoder across different target domains.

\begin{figure}[t]
\centering
    \includegraphics[width=0.98\linewidth]{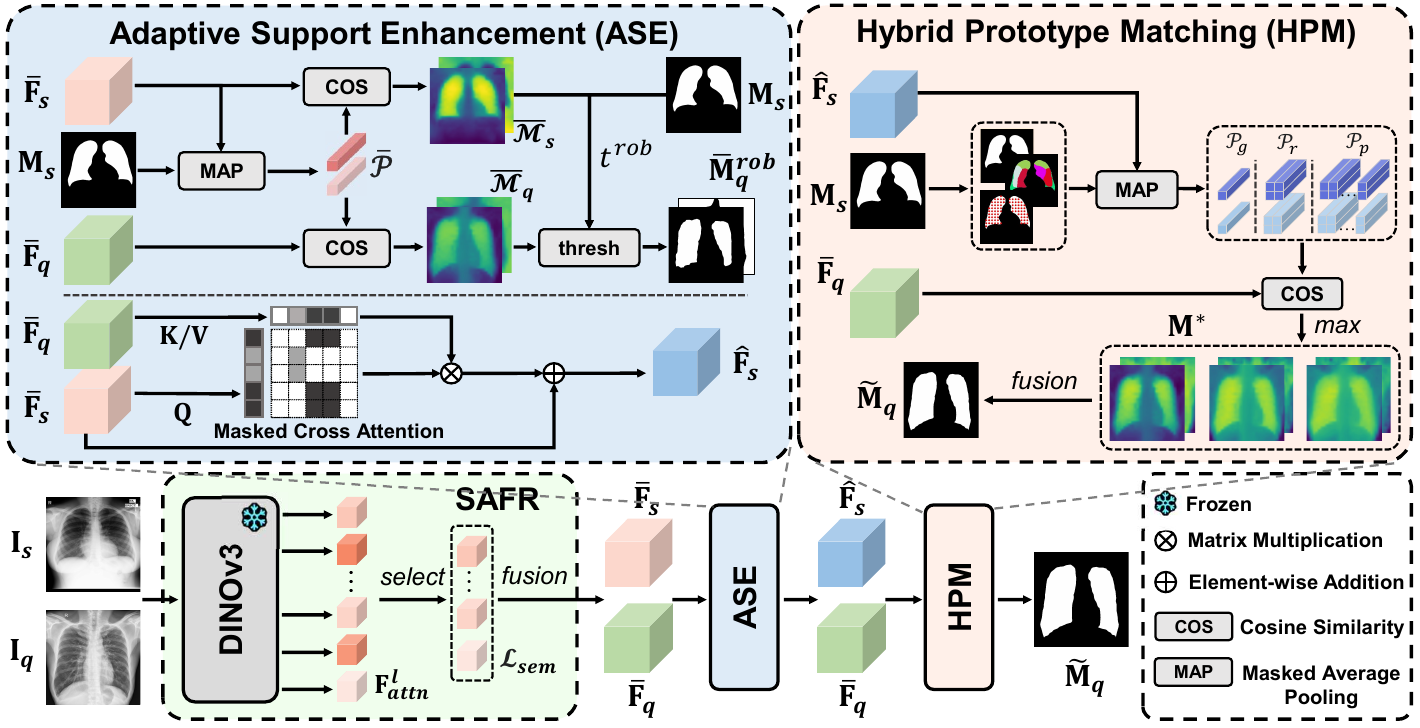}
    \vspace{-1ex}
    \caption{Overview of our method. We first utilize DINOv3 to extract multi-layer features of both support and query images. The SAFR then locates and re-fuses representative semantic-aware features from these extracted features. Subsequently, the fused support features are enhanced by the ASE module. Finally, the HPM computes and integrates the matching results between diverse support prototypes and query features to generate the final query prediction.}
    \vspace{-3ex}
    \label{fig:framework}
\end{figure}

\section{Methodology}
\label{sec:method}

\subsection{Problem Definition}
In the traditional CD-FSS task, the model is trained on the source domain $\mathcal{D}_s = \{X_s, Y_s\}$ and evaluated on the target domain $\mathcal{D}_t = \{X_t, Y_t\}$, where $X$ denotes the data distribution and $Y$ represents the corresponding label space. The source and target domains have no overlap in both data distributions and label spaces, \ie, $X_s \neq X_t$ and $Y_s \cap Y_t = \emptyset$.

Unlike traditional CD-FSS work, we directly generalize the model to different target domains during testing, without utilizing any source domain data for training. Each testing task is organized as an episode, consisting of a support set $\mathcal{S} = \{(\mathbf{I}^i_s, \mathbf{M}^i_s)\}_{i=1}^K$ and a query set $\mathcal{Q} = \{(\mathbf{I}_q, \mathbf{M}_q)\}$, where $K$ denotes the number of support samples, $\mathbf{I}$ denotes an image, and $\mathbf{M}$ denotes its binary mask. In each episode, the model predicts the query mask based on the support set $\mathcal{S}$ and the query image $\mathbf{I}_q$.

\subsection{Method Overview}
The overall architecture of our proposed training-free framework is illustrated in \cref{fig:framework}. First, support and query images from the target domain are fed into DINOv3 to extract multi-layer raw features. The Semantic-aware Feature Re-fusion (SAFR) adaptively identifies and re-fuses features that emphasize semantic patterns, thereby generating support and query features with enhanced semantic discriminability. After computing support prototypes, the Adaptive Support Enhancement (ASE) module constructs the similarity map between these prototypes and query features, generating robust query foreground and background region masks based on adaptive thresholds. Subsequently, the support features aggregate information from corresponding query regions through masked cross-attention, thereby improving semantic consistency between support and query features. Finally, the Hybrid Prototype Matching (HPM) module extracts global, regional, and pixel-level prototypes from enhanced support features and fuses the matching results between query features and these prototypes to generate the final query prediction. Note that all features refer to 2D feature maps obtained by spatially reshaping the local patch tokens.

\subsection{Semantic-aware Feature Re-fusion}
\begin{figure}[t]
\centering
    \includegraphics[width=0.97\linewidth]{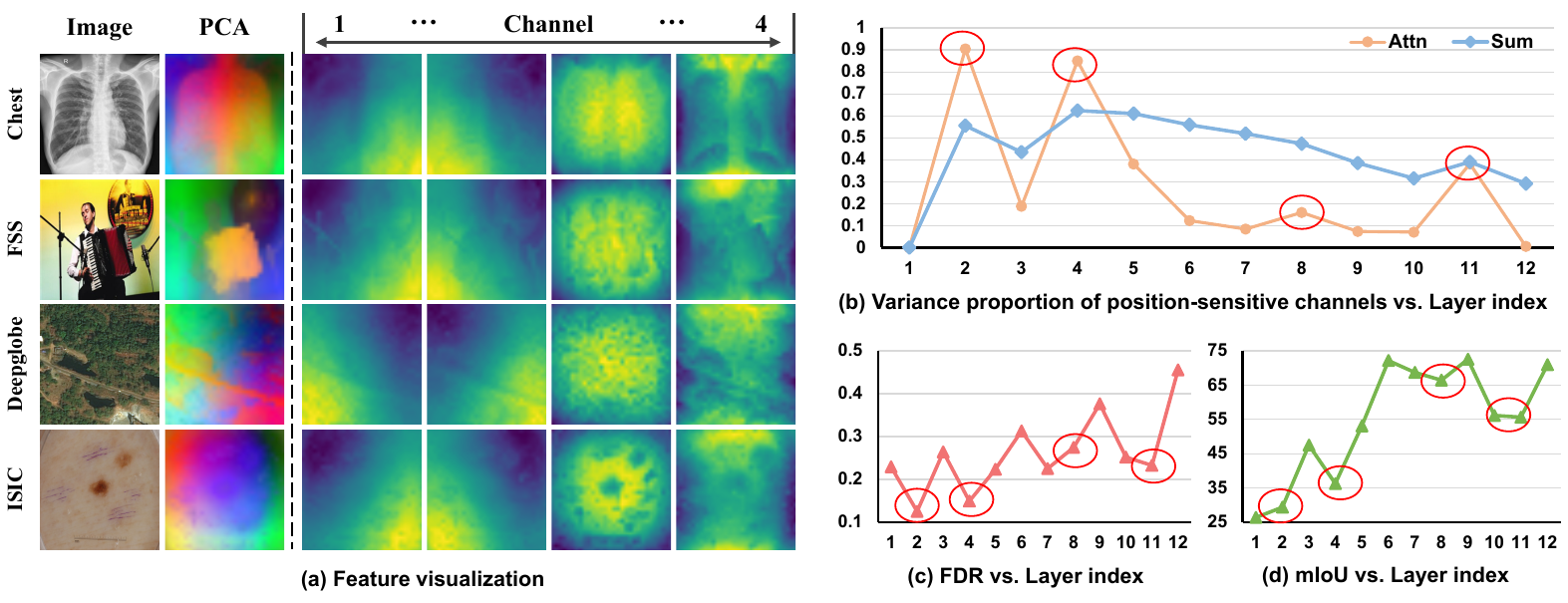}
    \vspace{-2ex}
    \caption{(a) We perform PCA on the last-layer feature $\mathbf{F}_{last}$ of DINOv3 and visualize the first three components, indicating that these features are overly focused on local consistency. Further visualization of the top four channels with the highest variance in $\mathbf{F}_{last}$ reveals that these channels typically encode patterns related to pixel positions rather than semantically meaningful concepts. (b)-(d) Statistical analysis of features across layers in DINOv3 on the Chest X-ray dataset.}
    \vspace{-3ex}
    \label{fig:feature_vis_analysis}
\end{figure}

\textbf{Analysis of DINOv3 Features.}
Extracting features with strong semantic discriminability is a key prerequisite for CD-FSS. To this end, given an input image $\mathbf{I}$, we first perform principal component analysis (PCA) on the last-layer feature $\mathbf{F}_{last}\in \mathbb{R}^{H\times W\times C}$ extracted by DINOv3 and visualize the top three components, where $H$, $W$, and $C$ denote the height, width, and channel depth of the feature, respectively. As shown in \cref{fig:feature_vis_analysis}(a), a notable observation is that these features exhibit strong local consistency, while their semantic discriminability is relatively weak. For further investigation, we extract and visualize the top four channels with the highest variance, finding that these channels typically encode patterns related to pixel positions rather than semantically meaningful concepts.

To trace the origin of these position-sensitive channels, we extract the raw outputs $\{\mathbf{F}^l_{attn}\}_{l=1}^L$ of attention modules and the fused outputs $\{\mathbf{F}^l_{sum}\}_{l=1}^L$ after residual connections in each block, where $L$ is the number of layers. Additionally, $\mathbf{F}_{sum}^l=\operatorname{FFN}(\operatorname{LN}(\mathbf{F}_{sum}^{l-1}+\mathbf{F}_{attn}^l))$ and $\mathbf{F}_{last}=\mathbf{F}_{sum}^L$, where $\operatorname{LN}$ denotes layer normalization, and $\operatorname{FFN}$ stands for a feed-forward network. Feature visualizations show that DINOv3 layers alternate between focusing on semantic patterns and positional variations. These position-aware features propagate via residual connections to $\mathbf{F}_{last}$, limiting its semantic discriminability. See the supplementary material for details.

For quantitative validation, we select the top $N_c=\operatorname{round}(C/\gamma)$ highest-variance channels from $F_{last}$ as the position-sensitive channel set $\mathcal{C}$, where $\gamma$ is a hyperparameter controlling the size of $\mathcal{C}$. Then, we compute the variance proportion $r$ of $\mathcal{C}$ in each feature $\mathbf{F}$:
\begin{equation}
    r=\frac{\sum_{c\in\mathcal{C}}\sigma^2(\mathbf{F}_{c})}{\sum_{c=1}^{C}\sigma^2(\mathbf{F}_{c})},
\end{equation}
where $\sigma(\mathbf{F}_{c})$ denotes the standard deviation of the $c$-th channel in feature $\mathbf{F}$. As shown in \cref{fig:feature_vis_analysis}(b), the position-aware feature $\mathbf{F}^l_{attn}$ with higher $r^l_{attn}$ increases $r^l_{sum}$ of the fused feature $\mathbf{F}^l_{sum}$, and this persists until the last layer.

Furthermore, for each $\mathbf{F}^l_{attn}$, we compute its Fisher discriminant ratio (FDR) \cite{kan2016multi} $f^l_{attn}$ and evaluate its CD-FSS performance. The FDR is defined as the trace ratio between the between-class and within-class scatter matrices, reflecting the inter-class discriminability of features. \cref{fig:feature_vis_analysis} shows that layers with higher variance proportions generally exhibit lower FDR, indicating weaker foreground-background discriminability, and these layers also demonstrate relatively inferior segmentation performance.

\noindent
\textbf{Feature Re-fusion.}
Position-aware features weaken the semantic discriminability of $\mathbf{F}_{last}$. As shown in \cref{tab:enhance_strategy}, directly modifying or discarding these features alters the feature distribution in subsequent layers, resulting in a further drop in segmentation performance. Therefore, instead of using the standard final output $\mathbf{F}_{last}$, we re-fuse semantically-aware features to construct a representation with enhanced semantic discriminability.

Specifically, given an input image $\mathbf{I}$, we first extract the raw outputs $\{\mathbf{F}^l_{attn}\}_{l=1}^L$ and the final output $\mathbf{F}_{last}$, computing the position-sensitive channel set $\mathcal{C}$ from $\mathbf{F}_{last}$. We then compute the variance proportions $\{r^l_{attn}\}_{l=1}^L$ and FDRs $\{f^l_{attn}\}_{l=1}^L$ for each layer. Subsequently, we identify all layers where $r_{attn}$ exceeds the average as position-aware layers and add them to the position-aware layer set $\mathcal{L}_{pos}$:
\begin{equation}
    \mathcal{L}_{pos}=\left\{l_{pos}^i\right\}_{i=1}^{|\mathcal{L}_{pos}|}=\left\{l\;\middle|\;r^{l}_{attn} > \frac{1}{L}\sum_{l^{\prime}=1}^Lr^{l^{\prime}}_{attn}\right\}.
\end{equation}
We also add $L+1$ to $\mathcal{L}_{pos}$ and sort it in ascending order. Next, we sequentially select the layer with the minimum $r_{attn}$ between adjacent position-aware layers to form the representative semantic-aware layer set $\mathcal{L}_{sem}$:
\begin{equation}
    \mathcal{L}_{sem}=\bigcup_{i=0}^{|\mathcal{L}_{pos}|-1}\left\{\arg\min_{l\in(l_{pos}^{i},l_{pos}^{i+1})}r_{attn}^l\right\}.
\end{equation}

Finally, all representative semantic-aware features are weighted to obtain the fused feature $\bar{\mathbf{F}}$:
\begin{equation}
    \bar{\mathbf{F}}=\sum_{l\in \mathcal{L}_{sem}}\omega^l\mathbf{F}^l_{attn},\quad \omega^l=\frac{f^l_{attn}/r^l_{attn}}{\sum_{l^{\prime}\in\mathcal{L}_{sem}}f^{l^{\prime}}_{attn}/r^{l^{\prime}}_{attn}},
\end{equation}
where layers with lower $r_{attn}$ and higher $f_{attn}$ are assigned higher weights. We utilize support features to compute $\mathcal{L}_{pos}$, $\mathcal{L}_{sem}$, and $\omega^l$, applying them to both support and query features.

\subsection{Adaptive Support Enhancement}
Due to variations in foreground instances or background categories, the semantic consistency between some support and query features is weak, leading to insufficient cross-image semantic discrimination. Some FSS methods \cite{peng2023hierarchical,fan2022self,xu2023self} mitigate this by introducing query-support feature interaction or fusion modules. However, these methods typically involve trainable parameters or fixed thresholds, which fail to generalize across domains. To address this, we aim to design a support enhancement module that is training-free and adaptable to different domains, thereby reducing the semantic gap between support and query. The core idea is to aggregate query information into the support features via cross-attention without projection layers, while restricting the aggregation to robust query foreground/background regions to prevent feature confusion.

Specifically, after obtaining the fused features $\bar{\mathbf{F}}_s$ and $\bar{\mathbf{F}}_q$, we compute the support prototypes $\bar{\mathcal{P}}=\{\bar{\mathbf{p}}_{f}, \bar{\mathbf{p}}_{b}\}$ via masked average pooling (MAP), further computing their cosine similarity with $\bar{\mathbf{F}}_q$ to obtain the query similarity map $\bar{\mathcal{M}}_{q}=\{\bar{\mathbf{M}}_{q,f}, \bar{\mathbf{M}}_{q,b}\}=\operatorname{softmax}(\operatorname{cosine}(\bar{\mathcal{P}}, \bar{\mathbf{F}}_q))$. The support similarity map $\bar{\mathcal{M}}_{s}=\{\bar{\mathbf{M}}_{s,f}, \bar{\mathbf{M}}_{s,b}\}$ is computed in the same way.

To locate robust query foreground regions, we first sample a set of thresholds $\mathcal{T}_{f}=\{t_{f}^i\}_{i=1}^{N_t}$ uniformly from the interval (0, 1), where $N_t$ is the number of thresholds. Each $t_{f}^i$ is utilized to binarize the support foreground similarity map $\bar{\mathbf{M}}_{s,f}$ to obtain the corresponding support prediction mask $\bar{\mathbf{M}}^i_s=\mathbb{I}(\bar{\mathbf{M}}_{s,f}>t_{f}^i)$, where $\mathbb{I}$ is the indicator function. Subsequently, we compute the mIoU between $\bar{\mathbf{M}}^i_s$ and the ground-truth mask $\mathbf{M}_s$, selecting the threshold with the highest mIoU as the optimal threshold $t_{f}^*$. Since the values of support-support similarity are typically higher than support-query similarity, we compute a more robust threshold $t_{f}^{rob}$ instead of directly using $t_{f}^*$:
\begin{equation}
    t_{f}^{rob}=\frac{1}{|\mathcal{T}_{f}^{rob}|}\sum_{j\in \mathcal{T}_{f}^{rob}} t_{f}^j, \quad \mathcal{T}_{f}^{rob}=\{t|t\in \mathcal{T}_{f},t>\alpha t_{f}^*\},
\end{equation}
where $\alpha=0.75$ is a hyperparameter. We then use $t_{f}^{rob}$ to binarize $\bar{\mathbf{M}}_{q,f}$, obtaining the robust query foreground mask $\bar{\mathbf{M}}_{q,f}^{rob}$. The robust query background mask $\bar{\mathbf{M}}_{q,b}^{rob}$ is obtained in the same way.

Next, we reshape $\bar{\mathbf{F}}_s$ into $\mathbf{Q}\in \mathbb{R}^{HW \times C}$ and $\bar{\mathbf{F}}_q$ into $\mathbf{K}\in \mathbb{R}^{HW \times C}$ and $\mathbf{V}\in \mathbb{R}^{HW \times C}$, aggregating query information into support features through training-free cross-attention. To prevent information confusion, we restrict support foreground/background regions to interact only with robust query foreground/background regions, which is achieved by adding a mask $\mathbf{M}_{sq}\in \mathbb{R}^{HW\times HW}$ to the attention matrix. $\mathbf{M}_{sq}$ is defined as:
\begin{equation}
    \mathbf{M}_{sq}(\mathbf{u}_s,\mathbf{u}_q) =
    \begin{cases}
    0, & \text{if} \left(\bar{\mathbf{M}}_{q,f}^{rob}(\mathbf{u}_q)\mathbf{M}_s(\mathbf{u}_s) = 1\right) \\
    & \lor \left( \bar{\mathbf{M}}_{q,b}^{rob}(\mathbf{u}_q)\mathbf{M}_{s,b}(\mathbf{u}_s) = 1\right), \\
    -\infty, & \text{otherwise}
    \end{cases}
\end{equation}
where $\mathbf{u}_s=(x_s,y_s)$ and $\mathbf{u}_q=(x_q,y_q)$ denote spatial coordinates, and $\mathbf{M}_{s,b}=1-\mathbf{M}_s$ is the support background mask. The query information aggregation is formulated as:
\begin{equation}
    \mathbf{A}_{sq}=\operatorname{softmax}(\frac{\mathbf{Q}\mathbf{K}^T}{\sqrt{C}}+\mathbf{M}_{sq})\mathbf{V}.
\end{equation}
Finally, we reshape $\mathbf{A}_{sq}$ into $\mathbf{F}_{sq}\in \mathbb{R}^{H\times W\times C}$ and fuse it with $\bar{\mathbf{F}}_s$ to obtain the enhanced feature $\hat{\mathbf{F}}_s=(\bar{\mathbf{F}}_s+\mathbf{F}_{sq})/2$.

\subsection{Hybrid Prototype Matching}
Images from different target domains exhibit varying semantic complexity, making it difficult for a single form of support-query matching to adapt universally. We propose to fuse matching results from global, regional, and pixel-level prototypes to generate the final query prediction.

Specifically, we first compute three types of support foreground prototype sets. The global foreground prototype $\mathbf{p}_{g,f}$ is obtained by applying MAP to $\hat{\mathbf{F}}_s$ and $\mathbf{M}_s$, forming the global prototype set $\mathcal{P}_{g,f}=\{\mathbf{p}_{g,f}\}$. Simultaneously, we treat each foreground feature in $\hat{\mathbf{F}}_s$ as a pixel-level foreground prototype, constructing the pixel-level prototype set $\mathcal{P}_{p,f}=\{\mathbf{p}_{p,f}^i\}_{i=1}^{N{s,f}}$, where $N_{s,f}$ denotes the number of foreground pixels in $\hat{\mathbf{F}}_s$. Additionally, we cluster the support foreground features into $N_r$ clusters using k-means++ \cite{arthur2006k} and regard their centers as regional prototypes, forming the regional prototype set $\mathcal{P}_{r,f}=\{\mathbf{p}_{r,f}^i\}_{i=1}^{N_r}$. Corresponding background prototype sets $\mathcal{P}_{g,b}$, $\mathcal{P}_{p,b}$, and $\mathcal{P}_{r,b}$ are obtained in a similar manner. The only difference is that, since the background typically contains more semantic categories, the support background features are clustered into $2N_r$ clusters to generate $\mathcal{P}_{r,b}$.

Next, we compute the similarity map for each prototype set. Specifically, given a prototype set $\mathcal{P}_{m,n}=\{\mathbf{p}^i_{m,n}\}_{i=1}^{|\mathcal{P}_{m,n}|}$, where $m \in \{g,r,p\}$ and $n \in \{f,b\}$, we compute the cosine similarity between each prototype $\mathbf{p}_{m,n}^i$ and $\bar{\mathbf{F}}_q$, obtaining a similarity map $\mathbf{M}_{m,n}^i \in \mathbb{R}^{H \times W}$. We then aggregate by taking the pixel-wise maximum to capture the strongest association between each pixel and any prototype, generating the similarity map $\mathbf{M}_{m,n}^*$:
\begin{equation}
    \mathbf{M}_{m,n}^*(x,y)=\max_{i\in\{1,\ldots,|\mathcal{P}_{m,n}|\}}\mathbf{M}_{m,n}^i(x,y).
\end{equation}

Subsequently, we compute the pseudo query mask for each prototype type as $\tilde{\mathbf{M}}_{m} = \mathbb{I}(\mathbf{M}^{*}_{m,f} > \mathbf{M}^{*}_{m,b})$, which is further utilized to compute the average foreground similarity $\mu_{m,f}$ and average background similarity $\mu_{m,b}$ from $\mathbf{M}^{*}_{m,f}$. The difference between these averages, $\omega_{m} = \mu_{m,f} - \mu_{m,b}$, is employed as both the confidence measure and fusion weight for prototype type $m$. After normalizing the weights across types with softmax, we compute the fused foreground and background similarity maps as:
\begin{equation}
    \tilde{\mathbf{M}}_f=\sum_{m} \omega_{m} \mathbf{M}^{*}_{m,f}, \quad \tilde{\mathbf{M}}_b=\sum_{m} \omega_{m} \mathbf{M}^{*}_{m,b}.
\end{equation}

Finally, the query prediction $\tilde{\mathbf{M}}_q$ is obtained by comparing the fused maps, \ie, $\tilde{\mathbf{M}}_q=\mathbb{I}(\tilde{\mathbf{M}}_f>\tilde{\mathbf{M}}_b)$.

\section{Experiments}
\label{sec:exper}

\subsection{Datasets and Implementation Details}
\textbf{Datasets.}
Following the setup of PATNet \cite{lei2022cross}, we evaluate our method on four target domain datasets: Deepglobe \cite{demir2018deepglobe}, ISIC \cite{codella2019skin,tschandl2018ham10000}, Chest X-ray \cite{candemir2013lung,jaeger2013automatic}, and FSS-1000 \cite{li2020fss}. Unlike existing CD-FSS methods, our approach does not leverage any additional source domain datasets (\eg, PASCAL \cite{everingham2010pascal} and COCO \cite{lin2014microsoft}) for training. Please refer to our supplementary material for more details.

\noindent
\textbf{Implementation Details.}
We employ DINOv3 ViT-B/16 \cite{simeoni2025dinov3} as our backbone to extract high-quality visual representations. Following LoEC \cite{liu2025devil}, we resize all images to $480 \times 480$. The hyperparameters $\gamma$, $\alpha$, $N_t$, and $N_r$ are set to 200, 0.75, 20, and 9, respectively. Consistent with previous methods, we use mean intersection over union (mIoU) as the evaluation metric and report the average results over 5 random seeds. Each run consists of 1200 episodes sampled from Deepglobe, ISIC, and Chest X-ray, respectively, and 2400 episodes from FSS-1000. All experiments are conducted on an NVIDIA GeForce RTX 4090 GPU.

\begin{table*}[t]
    \caption{Mean-IoU of 1-shot and 5-shot results compared with previous methods. The best and second-best methods are highlighted in \textbf{bold} and \underline{underlined}, respectively. $\dag$ marks methods implemented or reproduced by us.}
    \vspace{-1ex}
    \label{tab:performance}
    \centering
    \setlength{\tabcolsep}{4.8pt}
    \resizebox{\textwidth}{!}{
    \begin{tabular}{l|c|c|c|cc|cc|cc|cc|cc}
        \toprule
        \rule{0pt}{2.5ex}
        \multirow{2}*{Methods} & \multirow{2}*{Task} & \multirow{2}*{Mark} & \multirow{2}*{Backbone} & \multicolumn{2}{c|}{Deepglobe} & \multicolumn{2}{c|}{ISIC} & \multicolumn{2}{c|}{Chest X-ray} & \multicolumn{2}{c|}{FSS-1000} & \multicolumn{2}{c}{Average} \\
        \cline{5-14}
        \rule{0pt}{2.5ex}
        & & & & \multicolumn{1}{c}{1-shot} & 5-shot & \multicolumn{1}{c}{1-shot} & 5-shot & \multicolumn{1}{c}{1-shot} & 5-shot & \multicolumn{1}{c}{1-shot} & 5-shot & \multicolumn{1}{c}{1-shot} & 5-shot \\
        \midrule
        \multicolumn{13}{c}{Training-based Methods} \\
        \midrule
        SSP \cite{fan2022self} & FSS & ECCV-22 & ResNet-50 & 40.00 & 48.68 & 35.49 & 45.86 & 74.44 & 74.26 & 78.91 & 80.59 & 57.21 & 62.35 \\
        FPTrans \cite{zhang2022feature} & FSS & NIPS-22 & ViT-B & 38.36 & 49.30 & 48.65 & 60.37 & 80.92 & 82.91 & 80.74 & 83.65 & 62.17 & 69.06 \\
        HDMNet \cite{peng2023hierarchical} & FSS & CVPR-23 & ResNet-50 & 25.40 & 39.10 & 33.00 & 35.00 & 30.60 & 31.30 & 75.10 & 78.60 & 41.00 & 46.00 \\
        FSSAM$^{\dag}$ \cite{xu2025unlocking}  & FSS & ICML-25 & DINOv2-B+SAM2 & 40.25 & 46.58 & 47.75 & 59.68 & 77.11 & 86.56 & 80.36 & 87.11 & 61.37 & 69.98 \\
        PATNet \cite{lei2022cross} & CD-FSS & ECCV-22 & ResNet-50 & 37.89 & 42.97 & 41.16 & 53.58 & 66.61 & 70.20 & 78.59 & 81.23 & 56.06 & 61.99 \\
        PATNet$^{\dag}$ \cite{lei2022cross} & CD-FSS & ECCV-22 & DINOv3-B & 33.78 & 40.45 & \underline{54.37} & 62.36 & 62.69 & 64.42 & 83.71 & 84.65 & 58.64 & 62.97 \\
        ABCDFSS \cite{herzog2024adapt} & CD-FSS & CVPR-24 & ResNet-50 & 42.60 & 49.00 & 45.70 & 53.30 & 79.80 & 81.40 & 74.60 & 76.20 & 60.67 & 64.97 \\
        ABCDFSS$^{\dag}$ \cite{herzog2024adapt} & CD-FSS & CVPR-24 & DINOv3-B & 32.57 & 39.26 & 49.56 & 56.32 & 80.73 & 82.91 & 76.80 & 79.39 & 59.92 & 64.47 \\
        DRA \cite{su2024domain} & CD-FSS & CVPR-24 & ResNet-50 & 41.29 & 50.12 & 40.77 & 48.87 & 82.35 & 82.31 & 79.05 & 80.40 & 60.86 & 65.42 \\
        APSeg \cite{he2024apseg} & CD-FSS & CVPR-24 & SAM-B & 35.94 & 39.98 & 45.43 & 53.98 & \underline{84.10} & 84.50 & 79.71 & 81.90 & 61.30 & 65.09 \\
        APM \cite{tong2024lightweight} & CD-FSS & NIPS-24 & ResNet-50 & 40.86 & 44.92 & 41.71 & 51.16 & 78.25 & 82.81 & 79.29 & 81.83 & 60.03 & 65.18 \\
        LoEC \cite{liu2025devil} & CD-FSS & CVPR-25 & ViT-B & 42.12 & 51.48 & 52.91 & \underline{62.43} & 83.94 & 84.12 & 81.05 & 83.69 & \underline{65.01} & \underline{70.43} \\
        LoEC$^{\dag}$ \cite{liu2025devil} & CD-FSS & CVPR-25 & DINOv3-B & 44.34 & 53.25 & 50.98 & 61.38 & 73.83 & 74.58 & 84.33 & 86.16 & 63.37 & 68.84 \\
        SDRC \cite{tong2025self} & CD-FSS & ICML-25 & ViT-B & 43.15 & 46.83 & 46.57 & 55.02 & 82.86 & 84.79 & 80.31 & 82.55 & 63.22 & 67.30 \\
        DFN \cite{tong2025adapter} & CD-FSS & ICML-25 & ViT-B & 39.45 & 47.67 & 50.36 & 58.53 & 83.18 & \textbf{87.14} & 82.97 & 85.72 & 63.99 & 69.77 \\
        ISA \cite{fan2025adapting}  & CD-FSS & ICCV-25 & ResNet-50 & 44.30 & 52.70 & 37.20 & 56.10 & 83.40 & 86.30 & 78.80 & 86.00 & 60.90 & 70.30 \\
        \midrule
        \multicolumn{13}{c}{Training-free Methods} \\
        \midrule
        PerSAM \cite{zhang2023personalize}  & FSS & ICLR-24 & SAM-B & 36.08 & 40.65 & 23.27 & 25.33 & 29.95 & 30.05 & 60.92 & 66.53 & 37.56 & 40.64 \\
        Matcher$^{\dag}$ \cite{liu2023matcher}  & FSS & ICLR-24 & DINOv2-B+SAM-B & 47.01 & 50.98 & 38.29 & 39.22 & 76.42 & 79.04 & 85.27 & \textbf{88.52} & 61.75 & 64.44 \\
        Matcher$^{\dag}$ \cite{liu2023matcher}  & FSS & ICLR-24 & DINOv3-B+SAM-B & 46.85 & 53.73 & 38.68 & 46.85 & 75.86 & 81.07 & 83.39 & 87.44 & 61.20 & 67.27 \\
        GF-SAM$^{\dag}$ \cite{zhang2024bridge}  & FSS & NIPS-24 & DINOv2-B+SAM-B & \underline{49.05} & 55.13 & 44.47 & 50.25 & 38.70  & 37.93 & \underline{85.75} & 86.81 & 54.49 & 57.53 \\
        GF-SAM$^{\dag}$ \cite{zhang2024bridge}  & FSS & NIPS-24 & DINOv3-B+SAM-B & 48.75 & \underline{57.86} & 49.13 & 56.91 & 43.37 & 42.63 & \textbf{86.51} & \underline{87.57} & 56.94 & 61.24 \\
        MAUP$^{\dag}$ \cite{zhu2025maup}  & FSMIS & MICCAI-25 & DINOv2-B+SAM-B & 41.50 & 45.06 & 33.88 & 36.35 & 73.96 & 75.63 & 73.29 & 75.97 & 55.66 & 58.25 \\
        \midrule
        \textbf{Ours}  & CD-FSS & Ours & DINOv3-B & \textbf{49.71} & \textbf{58.88} & \textbf{55.73} & \textbf{62.55} & \textbf{85.44} & \underline{87.01} & 82.67 & 83.89 & \textbf{68.39} & \textbf{73.08} \\
        \bottomrule
    \end{tabular}}
    \vspace{-2ex}
\end{table*}

\subsection{Comparison with State-of-the-art Methods}
In \cref{tab:performance}, we compare our method with existing methods, including those using ResNet-50 \cite{he2016deep} and ViT-B/16 \cite{dosovitskiy2020image} with ImageNet \cite{russakovsky2015imagenet} pretrained weights as backbones, as well as methods employing vision foundation models (VFMs) such as DINOv2 \cite{oquab2023dinov2} and SAM \cite{kirillov2023segment} as backbones. For fair comparison, we also replace the backbone of several CD-FSS and training-free methods with DINOv3-B/16. The results demonstrate that our method achieves significant improvements under both 1-shot and 5-shot settings, surpassing all existing training-based and training-free methods without any training. Specifically, our method surpasses the state-of-the-art training-based method LoEC \cite{liu2025devil} by 3.38$\%$ and 2.65$\%$, and exceeds the state-of-the-art training-free method Matcher \cite{liu2023matcher} by 7.19$\%$ and 5.81$\%$ under 1-shot and 5-shot settings, respectively.

Notably, introducing DINOv3 into training-based methods yields marginal gains or even performance degradation, and other VFM-based methods \cite{xu2025unlocking, he2024apseg} also underperform compared to some VFM-free methods \cite{zhang2022feature, liu2025devil}. This indicates that training-based methods still suffer from the risk of overfitting. Existing training-free methods also exhibit suboptimal performance on CD-FSS, which we attribute to two primary factors: 1) The modules or parameters in these methods are carefully designed for single-domain settings, limiting their cross-domain generalization. 2) While these methods benefit from SAM's strong semantic discrimination on natural images (\eg, FSS-1000), this ability diminishes on domains with large gaps, failing to compensate for its inherently limited cross-image matching capability.

\subsection{Ablation Studies}
\begin{figure}[t]
\centering
    \includegraphics[width=0.97\linewidth]{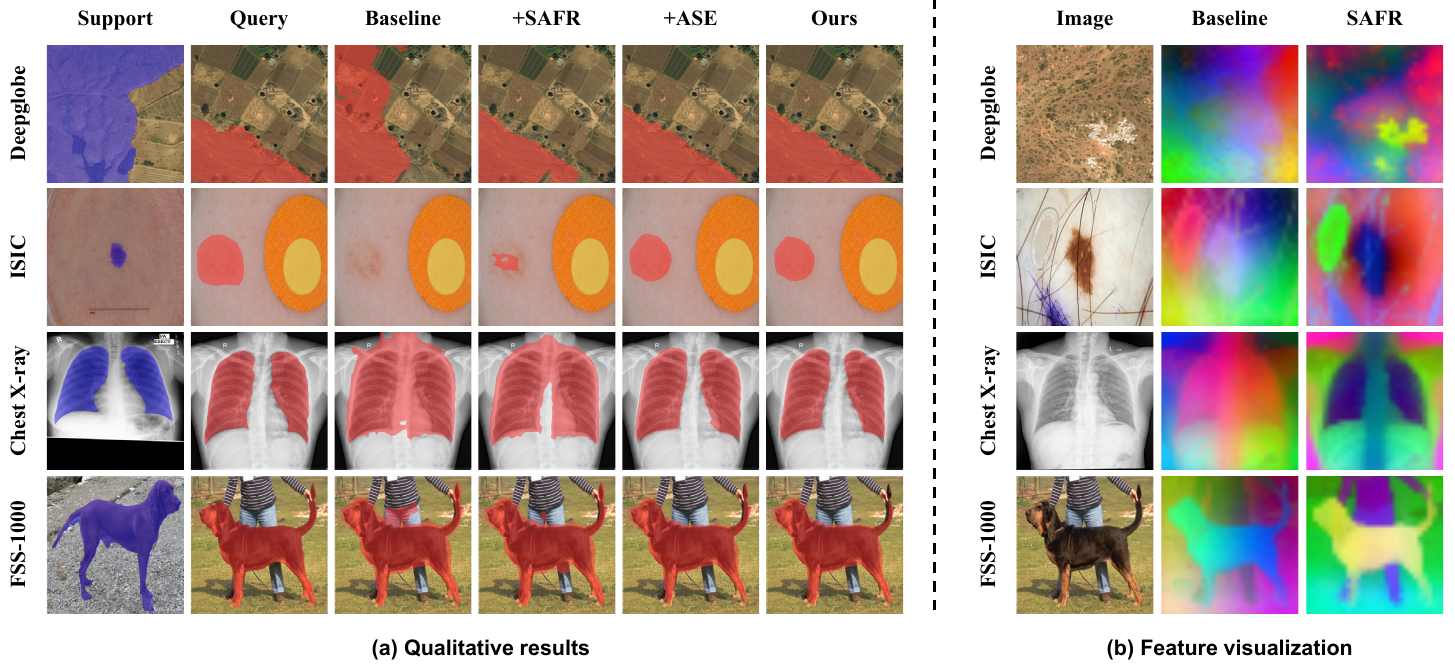}
    \vspace{-1ex}
    \caption{(a) Qualitative results with progressive module integration in 1-way 1-shot setting. (b) Visualizations of the first three PCA components demonstrate that SAFR generates representations with enhanced semantic discriminability.}
    \vspace{-1ex}
    \label{fig:vis_res_pca}
\end{figure}

\textbf{Effects of Each Component.}
We conduct ablation studies on each proposed component, with results shown in \cref{tab:component}. SAFR generates features with enhanced semantic discriminability by re-fusing representative semantic-aware features, improving the average mIoU by 3.79$\%$ and 4.10$\%$ under 1-shot and 5-shot settings, respectively. Building on this, ASE aggregates robust query information into the support features, enhancing semantic consistency between support and query features, and further improves the average mIoU by 3.63$\%$ and 2.22$\%$. HPM integrates matching results from diverse prototypes, resulting in additional mIoU gains of 3.46$\%$ and 2.97$\%$ for 1-shot and 5-shot settings, respectively. Finally, the combination of all three modules leads to further performance improvements. We also present qualitative results with progressive module integration in \cref{fig:vis_res_pca}(a). These results clearly demonstrate the effectiveness of each component.

\noindent
\textbf{Semantic Discriminability Enhancement Strategies.}
We compare the performance of different semantic discriminability enhancement strategies under the 1-shot setting, as shown in \cref{tab:enhance_strategy}. We first directly modify position-aware features, such as reducing their weight in residual connections or decreasing the variance proportion of position-sensitive channels through variance regularization. Since these strategies alter the feature distribution of subsequent layers, they actually lead to further degradation in performance. In contrast, explicitly removing the position-sensitive channels from the standard final output $\mathbf{F}_{last}$ results in a significant performance improvement, indicating that these channels suppress semantic discriminability. Our SAFR module avoids using $\mathbf{F}_{last}$ affected by position-sensitive channels, achieving better performance.

\begin{table}[t]
    \begin{minipage}{0.4\linewidth}
        \caption{Ablation studies for each proposed component.}
        \vspace{-1ex}
        \label{tab:component}
        \centering
        \setlength{\tabcolsep}{2.6mm}
        \resizebox{\linewidth}{!}{
        \begin{tabular}{ccc|cc}
            \toprule
            SAFR & ASE & HPM & 1-shot & 5-shot\\
            \midrule
            $\times$ & $\times$ & $\times$ & 58.74 & 64.54 \\
            \checkmark & $\times$ & $\times$ & 62.53 & 68.64 \\
            \checkmark & \checkmark & $\times$ & 66.16 & 70.86 \\
            \checkmark & $\times$ & \checkmark & 65.99 & 71.61 \\
            \checkmark & \checkmark & \checkmark & \textbf{68.39} & \textbf{73.08} \\ 
            \bottomrule
        \end{tabular}}
    \end{minipage}
    \hfill
    \begin{minipage}{0.55\linewidth}
        \caption{Ablation studies for different semantic discriminability enhancement strategies.}
        \vspace{-1ex}
        \label{tab:enhance_strategy}
        \centering
        \resizebox{\linewidth}{!}{
        \setlength{\tabcolsep}{4.8pt}
            \begin{tabular}{c|ccccc}
                \toprule
                Strategy & Deepglobe & ISIC & Chest & FSS & Average\\
                \midrule
                Baseline & 42.32 & 51.15 & 67.18 & 74.32 & 58.74 \\
                Reduce weight & 40.75 & 49.64 & 70.20 & 72.53 & 58.28 \\
                Reduce variance & 41.40 & 50.54 & 65.06 & 73.74 & 57.69 \\
                Remove & 48.34 & 53.26 & 71.42 & 75.58 & 62.15 \\
                SAFR & 48.78 & 53.69 & 72.03 & 75.60 & \textbf{62.53} \\
                \bottomrule
            \end{tabular}}
    \end{minipage}
    \vspace{-2ex}
\end{table}

\noindent
\textbf{Effects of Threshold Types.}
To validate the effectiveness of the robust threshold in ASE, we compare the performance of different threshold types under 1-shot setting, as summarized in \cref{tab:thresh_type}. First, the attention matrix without adding a mask may lead to information confusion, resulting in limited performance improvement. Simply using a fixed threshold of 0.5 to localize coarse foreground and background regions in the query image achieves further gains. Employing the proposed robust threshold better adapts to different target domains, achieving the best performance.

\begin{table}[t]
    \begin{minipage}{0.5\linewidth}
        \caption{Ablation studies for the effects of threshold types.}
        \vspace{-1ex}
        \label{tab:thresh_type}
        \centering
        \setlength{\tabcolsep}{1.2mm}
        \resizebox{\linewidth}{!}{
            \begin{tabular}{c|ccccc}
                \toprule
                & Deepglobe & ISIC & Chest & FSS & Average\\
                \midrule
                w/o mask & 49.52 & 54.57 & 73.16 & 77.80 & 63.76 \\
                Fixed threshold & 49.64 & 56.03 & 74.34 & 78.11 & 64.53 \\
                Best threshold & 48.67 & 55.47 & 80.03 & 79.62 & 65.95 \\
                Robust threshold & 49.48 & 55.98 & 80.14 & 79.03 & \textbf{66.16} \\
                \bottomrule
            \end{tabular}}
    \end{minipage}
    \hfill
    \begin{minipage}{0.45\linewidth}
        \caption{Ablation studies for the effects of prototype types.}
        \vspace{-1ex}
        \label{tab:proto_type}
        \centering
        \setlength{\tabcolsep}{1.5mm}
        \resizebox{\linewidth}{!}{    
            \begin{tabular}{c|ccccc}
                \toprule
                Prototype& Deepglobe & ISIC & Chest & FSS & Average\\
                \midrule
                Global & 49.48 & 55.98 & 80.14 & 79.03 & 66.16 \\
                Regional & 47.98 & 54.06 & 85.48 & 82.80 & 67.58 \\
                Pixel & 47.65 & 51.03 & 84.55 & 83.64 & 66.72 \\
                HPM & 49.71 & 55.73 & 85.44 & 82.67 & \textbf{68.39} \\
                \bottomrule
            \end{tabular}}
    \end{minipage}
\end{table}

\noindent
\textbf{Effects of Prototype Types.}
As shown in \cref{tab:proto_type}, we conduct ablation studies under the 1-shot setting to validate the necessity of diverse prototypes. Specifically, we first extract and match only a single type of prototype, and the results indicate that different domains favor different prototype types. By integrating the matching results of diverse prototypes, our method achieves a more balanced and improved performance across different target domains.

More ablation studies can be found in the supplementary material.

\begin{figure}[t]
    \begin{minipage}{0.48\linewidth}
        \setlength{\belowcaptionskip}{8pt}
        \setlength{\abovecaptionskip}{0pt}
        \captionof{table}{Results of combination with VFM-based methods.}
        \vspace{-1ex}
        \label{tab:combination}
        \centering
        \setlength{\tabcolsep}{1.5mm}
        \resizebox{\linewidth}{!}{    
            \begin{tabular}{c|ccccc}
                \toprule
                & Deepglobe & ISIC & Chest & FSS & Average\\
                \midrule
                GF-SAM \cite{zhang2024bridge} & 49.05 & 44.47 & 38.70 & 85.75 & 54.49 \\
                GF-SAM + Ours & 47.26 & 48.24 & 61.46 & 85.98 & \textbf{60.74} \\
                \midrule
                FSSAM \cite{xu2025unlocking} & 40.25 & 47.75 & 77.11 & 80.36 & 61.37 \\
                FSSAM + Ours & 47.67 & 43.59 & 84.36 & 85.23 & \textbf{65.21} \\
                \midrule
                Ours & 49.71 & 55.73 & 85.44 & 82.67 & \textbf{68.39} \\
                \bottomrule
            \end{tabular}}\textsuperscript{(\Letter)}
    \end{minipage}
    \hfill
    \begin{minipage}{0.48\linewidth}
        \centering
        \includegraphics[width=0.90\linewidth]{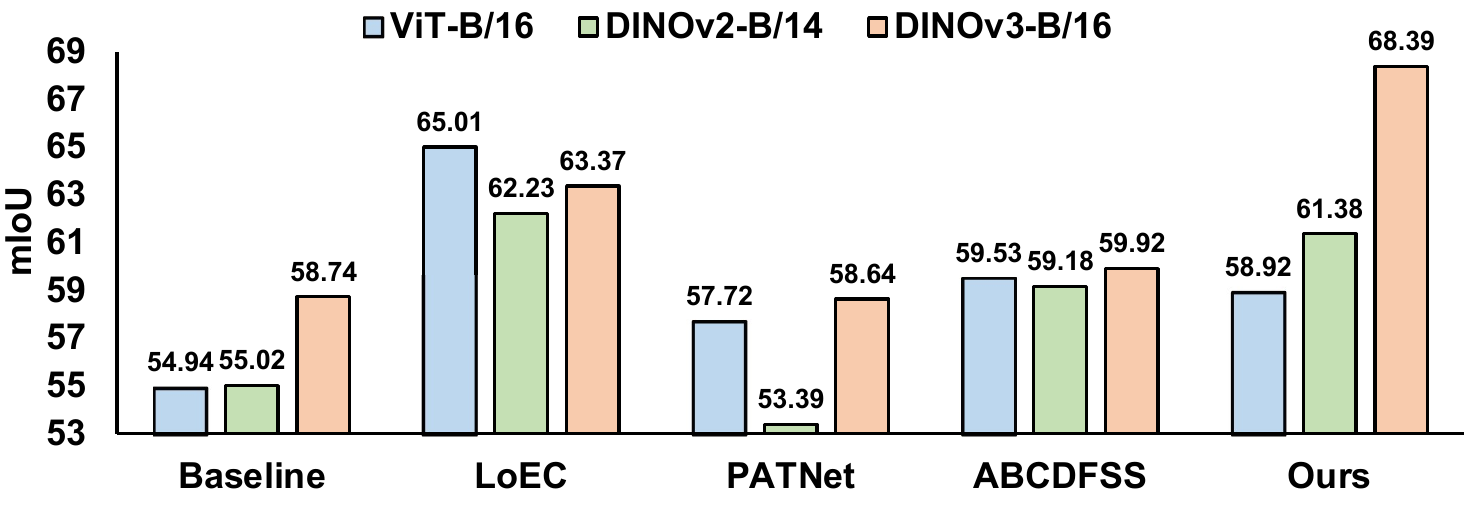}
        \caption{Performance of CD-FSS methods with different pretrained models.}
        \vspace{-1ex}
        \label{fig:pretrained}
    \end{minipage}
    \vspace{-1ex}
\end{figure}

\subsection{More Analysis}
\textbf{Combination with VFM-based Methods.}
Existing VFM-based methods \cite{xu2025unlocking, zhang2024bridge} typically leverage features from models like DINOv2 \cite{oquab2023dinov2} to compute support-query similarity as a prior for SAM \cite{kirillov2023segment, ravi2024sam}, reformulating cross-image matching segmentation into intra-image semantic partitioning at which SAM excels. The similarity maps computed by our method can also naturally serve as priors to combine with these methods. As shown in \cref{tab:combination}, our method significantly improves the performance of existing VFM-based methods on CD-FSS. However, the combined performance is inferior to that using our original priors alone, revealing the limitations of SAM in cross-domain scenarios.

\noindent
\textbf{Effects of pretrained models on CD-FSS.}
We incorporate DINOv2 \cite{oquab2023dinov2} and DINOv3 \cite{simeoni2025dinov3} into three different CD-FSS paradigms to explore the impact of stronger pretrained models on CD-FSS, as shown in \cref{fig:pretrained}. The baseline results approximately reflect the potential capability of models. For LoEC \cite{liu2025devil}, which treats the backbone as trainable parameters, performance decreases after introducing VFMs because the rich prior accelerates overfitting. Methods freezing the pretrained model (PATNet \cite{lei2022cross} and ABCDFSS \cite{herzog2024adapt}) degrade with DINOv2 but improve slightly with DINOv3. We argue that VFMs still accelerate overfitting of trainable parameters in these methods while providing better feature representations during inference. When VFM priors are sufficiently strong, the benefits during inference outweigh the overfitting effects in training, leading to marginal improvements. In contrast, our training-free method completely avoids overfitting risks and fully exploits the well-learned knowledge in VFMs.

\noindent
\textbf{Semantic Discriminability.}
Semantic discriminability of features is crucial for effective foreground-background segmentation. As shown in \cref{fig:vis_res_pca}(b), we perform PCA on the standard final output $\mathbf{F}_{last}$ and the fused feature $\bar{\mathbf{F}}$ obtained by SAFR, visualizing the first three components. The results demonstrate that our SAFR generates representations with enhanced semantic discriminability. Furthermore, cross-image semantic discriminability has a more direct impact on CD-FSS performance. Therefore, we compute the Fisher discriminant ratio between query foreground and support background, as well as between query foreground and support foreground, and use the difference between the two to measure cross-image semantic discriminability. As shown in \cref{tab:discriminability}, both SAFR and ASE effectively enhance the cross-image semantic discriminability.

\begin{table}[t]
    \begin{minipage}{0.475\linewidth}
        \caption{Effects of different modules on cross-image semantic discriminability.}
        \label{tab:discriminability}
        \centering
        \setlength{\tabcolsep}{1.5mm}
        \resizebox{\linewidth}{!}{    
            \begin{tabular}{c|cccc}
                \toprule
                & Deepglobe & ISIC & Chest & FSS\\
                \midrule
                Baseline & -0.0082 & 0.2471 & 0.2381 & 0.2724 \\
                Baseline+SAFR & 0.0848 & 0.4112 & 0.2750 & 0.4130 \\
                Baseline+SAFR+ASE & 0.4474 & 1.1880 & 0.5941 & 1.4488 \\
                \bottomrule
            \end{tabular}}
    \end{minipage}
    \hfill
    \begin{minipage}{0.475\linewidth}
        \caption{Model efficiency and performance under 1-shot setting.}
        \label{tab:efficiency}
        \centering   
        \setlength{\tabcolsep}{1.5mm}
        \resizebox{\linewidth}{!}{ 
            \begin{tabular}{c|cccc}
                \toprule
                Method &T. Params (M) & FLOPs (G) & FPS & mean-IoU\\
                \midrule
                LoEC \cite{liu2025devil} & 72.82 & 150.1 & 52.91 & 65.01 \\
                Matcher \cite{liu2023matcher} & 0 & 727.6 & 0.47 & 61.75 \\
                Ours & 0 & 161.9 & 47.62 & 68.39 \\
                \bottomrule
            \end{tabular}
        }
    \end{minipage}
    \vspace{-1ex}
\end{table}

\noindent
\textbf{Model Efficiency.}
We compare the model efficiency with the training-based state-of-the-art LoEC \cite{liu2025devil} and the training-free state-of-the-art Matcher \cite{liu2023matcher}, including the number of trainable parameters (T. Params), computational complexity (FLOPs), and inference speed (FPS), as shown in \cref{tab:efficiency}. Compared to LoEC, our model requires no additional training cost and achieves significantly better performance with comparable inference speed. Although Matcher is training-free, its complex multi-model architecture and pipeline result in prohibitive inference speed and inferior performance.

\section{Conclusion}
\label{sec:conclusion}
In this paper, we propose a training-free framework for CD-FSS to avoid the training overhead and overfitting risks, which addresses CD-FSS challenges by enhancing feature semantic discriminability and performing robust matching. Specifically, we propose an SAFR module to identify and re-fuse semantic-aware features, generating representations with enhanced semantic discriminability. To improve semantic consistency between support and query features, we propose an ASE module that adaptively locates robust query regions and aggregates query information from corresponding regions for support features. We also propose an HPM module that adapts to semantic complexity variations across different domains by integrating matching results from diverse prototypes. Extensive experiments demonstrate that our method achieves state-of-the-art performance on four datasets with different domain gaps without any training.

\section*{Acknowledgments}
This work was supported in part by the National Natural Science Foundation of China under the Grants No. 62371235 and No. U25A20444, in part by the Key Research and Development Plan of Jiangsu Province under Grant No. BE2023008-2.

%
%
\bibliographystyle{splncs04}
\bibliography{main}

@String(AAAI  = {AAAI})

@inproceedings{cheng2022masked,
  title={Masked-attention mask transformer for universal image segmentation},
  author={Cheng, Bowen and Misra, Ishan and Schwing, Alexander G and Kirillov, Alexander and Girdhar, Rohit},
  booktitle={Proceedings of the IEEE/CVF conference on computer vision and pattern recognition},
  pages={1290--1299},
  year={2022}
}

@inproceedings{yu2024embedding,
  title={Embedding-free transformer with inference spatial reduction for efficient semantic segmentation},
  author={Yu, Hyunwoo and Cho, Yubin and Kang, Beoungwoo and Moon, Seunghun and Kong, Kyeongbo and Kang, Suk-Ju},
  booktitle={European Conference on Computer Vision},
  pages={92--110},
  year={2024},
  organization={Springer}
}

@inproceedings{fu2025segman,
  title={SegMAN: Omni-scale context modeling with state space models and local attention for semantic segmentation},
  author={Fu, Yunxiang and Lou, Meng and Yu, Yizhou},
  booktitle={Proceedings of the Computer Vision and Pattern Recognition Conference},
  pages={19077--19087},
  year={2025}
}

@inproceedings{sun2024vrp,
  title={VRP-SAM: SAM with visual reference prompt},
  author={Sun, Yanpeng and Chen, Jiahui and Zhang, Shan and Zhang, Xinyu and Chen, Qiang and Zhang, Gang and Ding, Errui and Wang, Jingdong and Li, Zechao},
  booktitle={Proceedings of the IEEE/CVF Conference on Computer Vision and Pattern Recognition},
  pages={23565--23574},
  year={2024}
}

@inproceedings{peng2023hierarchical,
  title={Hierarchical dense correlation distillation for few-shot segmentation},
  author={Peng, Bohao and Tian, Zhuotao and Wu, Xiaoyang and Wang, Chengyao and Liu, Shu and Su, Jingyong and Jia, Jiaya},
  booktitle={Proceedings of the IEEE/CVF conference on computer vision and pattern recognition},
  pages={23641--23651},
  year={2023}
}

@inproceedings{fan2022self,
  title={Self-support few-shot semantic segmentation},
  author={Fan, Qi and Pei, Wenjie and Tai, Yu-Wing and Tang, Chi-Keung},
  booktitle={European conference on computer vision},
  pages={701--719},
  year={2022},
  organization={Springer}
}

@inproceedings{wang2024rethinking,
  title={Rethinking prior information generation with clip for few-shot segmentation},
  author={Wang, Jin and Zhang, Bingfeng and Pang, Jian and Chen, Honglong and Liu, Weifeng},
  booktitle={Proceedings of the IEEE/CVF conference on computer vision and pattern recognition},
  pages={3941--3951},
  year={2024}
}

@inproceedings{lei2022cross,
  title={Cross-domain few-shot semantic segmentation},
  author={Lei, Shuo and Zhang, Xuchao and He, Jianfeng and Chen, Fanglan and Du, Bowen and Lu, Chang-Tien},
  booktitle={European conference on computer vision},
  pages={73--90},
  year={2022},
  organization={Springer}
}

@inproceedings{su2024domain,
  title={Domain-rectifying adapter for cross-domain few-shot segmentation},
  author={Su, Jiapeng and Fan, Qi and Pei, Wenjie and Lu, Guangming and Chen, Fanglin},
  booktitle={Proceedings of the IEEE/CVF conference on Computer Vision and Pattern Recognition},
  pages={24036--24045},
  year={2024}
}

@inproceedings{nie2024cross,
  title={Cross-domain few-shot segmentation via iterative support-query correspondence mining},
  author={Nie, Jiahao and Xing, Yun and Zhang, Gongjie and Yan, Pei and Xiao, Aoran and Tan, Yap-Peng and Kot, Alex C and Lu, Shijian},
  booktitle={Proceedings of the IEEE/CVF Conference on Computer Vision and Pattern Recognition},
  pages={3380--3390},
  year={2024}
}

@inproceedings{liu2025devil,
  title={The devil is in low-level features for cross-domain few-shot segmentation},
  author={Liu, Yuhan and Zou, Yixiong and Li, Yuhua and Li, Ruixuan},
  booktitle={Proceedings of the Computer Vision and Pattern Recognition Conference},
  pages={4618--4627},
  year={2025}
}

@article{tong2025self,
  title={Self-Disentanglement and Re-Composition for Cross-Domain Few-Shot Segmentation},
  author={Tong, Jintao and Zou, Yixiong and Chen, Guangyao and Li, Yuhua and Li, Ruixuan},
  journal={arXiv preprint arXiv:2506.02677},
  year={2025}
}

@article{tong2024lightweight,
  title={Lightweight frequency masker for cross-domain few-shot semantic segmentation},
  author={Tong, Jintao and Zou, Yixiong and Li, Yuhua and Li, Ruixuan},
  journal={Advances in Neural Information Processing Systems},
  volume={37},
  pages={96728--96749},
  year={2024}
}

@article{tong2025adapter,
  title={Adapter Naturally Serves as Decoupler for Cross-Domain Few-Shot Semantic Segmentation},
  author={Tong, Jintao and Ma, Ran and Zou, Yixiong and Chen, Guangyao and Li, Yuhua and Li, Ruixuan},
  journal={arXiv preprint arXiv:2506.07376},
  year={2025}
}

@inproceedings{herzog2024adapt,
  title={Adapt before comparison: A new perspective on cross-domain few-shot segmentation},
  author={Herzog, Jonas},
  booktitle={Proceedings of the IEEE/CVF conference on computer vision and pattern recognition},
  pages={23605--23615},
  year={2024}
}

@article{russakovsky2015imagenet,
  title={Imagenet large scale visual recognition challenge},
  author={Russakovsky, Olga and Deng, Jia and Su, Hao and Krause, Jonathan and Satheesh, Sanjeev and Ma, Sean and Huang, Zhiheng and Karpathy, Andrej and Khosla, Aditya and Bernstein, Michael and others},
  journal={International journal of computer vision},
  volume={115},
  number={3},
  pages={211--252},
  year={2015},
  publisher={Springer}
}

@inproceedings{caron2021emerging,
  title={Emerging properties in self-supervised vision transformers},
  author={Caron, Mathilde and Touvron, Hugo and Misra, Ishan and J{\'e}gou, Herv{\'e} and Mairal, Julien and Bojanowski, Piotr and Joulin, Armand},
  booktitle={Proceedings of the IEEE/CVF international conference on computer vision},
  pages={9650--9660},
  year={2021}
}

@article{oquab2023dinov2,
  title={Dinov2: Learning robust visual features without supervision},
  author={Oquab, Maxime and Darcet, Timoth{\'e}e and Moutakanni, Th{\'e}o and Vo, Huy and Szafraniec, Marc and Khalidov, Vasil and Fernandez, Pierre and Haziza, Daniel and Massa, Francisco and El-Nouby, Alaaeldin and others},
  journal={arXiv preprint arXiv:2304.07193},
  year={2023}
}

@inproceedings{kirillov2023segment,
  title={Segment anything},
  author={Kirillov, Alexander and Mintun, Eric and Ravi, Nikhila and Mao, Hanzi and Rolland, Chloe and Gustafson, Laura and Xiao, Tete and Whitehead, Spencer and Berg, Alexander C and Lo, Wan-Yen and others},
  booktitle={Proceedings of the IEEE/CVF international conference on computer vision},
  pages={4015--4026},
  year={2023}
}

@inproceedings{jose2025dinov2,
  title={Dinov2 meets text: A unified framework for image-and pixel-level vision-language alignment},
  author={Jose, Cijo and Moutakanni, Th{\'e}o and Kang, Dahyun and Baldassarre, Federico and Darcet, Timoth{\'e}e and Xu, Hu and Li, Daniel and Szafraniec, Marc and Ramamonjisoa, Micha{\"e}l and Oquab, Maxime and others},
  booktitle={Proceedings of the Computer Vision and Pattern Recognition Conference},
  pages={24905--24916},
  year={2025}
}

@inproceedings{yang2024depth,
  title={Depth anything: Unleashing the power of large-scale unlabeled data},
  author={Yang, Lihe and Kang, Bingyi and Huang, Zilong and Xu, Xiaogang and Feng, Jiashi and Zhao, Hengshuang},
  booktitle={Proceedings of the IEEE/CVF conference on computer vision and pattern recognition},
  pages={10371--10381},
  year={2024}
}

@article{zhang2024bridge,
  title={Bridge the points: Graph-based few-shot segment anything semantically},
  author={Zhang, Anqi and Gao, Guangyu and Jiao, Jianbo and Liu, Chi and Wei, Yunchao},
  journal={Advances in Neural Information Processing Systems},
  volume={37},
  pages={33232--33261},
  year={2024}
}

@article{simeoni2025dinov3,
  title={Dinov3},
  author={Sim{\'e}oni, Oriane and Vo, Huy V and Seitzer, Maximilian and Baldassarre, Federico and Oquab, Maxime and Jose, Cijo and Khalidov, Vasil and Szafraniec, Marc and Yi, Seungeun and Ramamonjisoa, Micha{\"e}l and others},
  journal={arXiv preprint arXiv:2508.10104},
  year={2025}
}

@article{yuan2025ad,
  title={AD-DINOv3: Enhancing DINOv3 for Zero-Shot Anomaly Detection with Anomaly-Aware Calibration},
  author={Yuan, Jingyi and Ye, Jianxiong and Chen, Wenkang and Gao, Chenqiang},
  journal={arXiv preprint arXiv:2509.14084},
  year={2025}
}

@article{li2025meddinov3,
  title={MedDINOv3: How to adapt vision foundation models for medical image segmentation?},
  author={Li, Yuheng and Wu, Yizhou and Lai, Yuxiang and Hu, Mingzhe and Yang, Xiaofeng},
  journal={arXiv preprint arXiv:2509.02379},
  year={2025}
}

@article{yang2025segdino,
  title={Segdino: An efficient design for medical and natural image segmentation with dino-v3},
  author={Yang, Sicheng and Wang, Hongqiu and Xing, Zhaohu and Chen, Sixiang and Zhu, Lei},
  journal={arXiv preprint arXiv:2509.00833},
  year={2025}
}

@inproceedings{zhang2019canet,
  title={Canet: Class-agnostic segmentation networks with iterative refinement and attentive few-shot learning},
  author={Zhang, Chi and Lin, Guosheng and Liu, Fayao and Yao, Rui and Shen, Chunhua},
  booktitle={Proceedings of the IEEE/CVF conference on computer vision and pattern recognition},
  pages={5217--5226},
  year={2019}
}

@inproceedings{yang2020prototype,
  title={Prototype mixture models for few-shot semantic segmentation},
  author={Yang, Boyu and Liu, Chang and Li, Bohao and Jiao, Jianbin and Ye, Qixiang},
  booktitle={Computer Vision--ECCV 2020: 16th European Conference, Glasgow, UK, August 23--28, 2020, Proceedings, Part VIII 16},
  pages={763--778},
  year={2020},
  organization={Springer}
}

@inproceedings{li2021adaptive,
  title={Adaptive prototype learning and allocation for few-shot segmentation},
  author={Li, Gen and Jampani, Varun and Sevilla-Lara, Laura and Sun, Deqing and Kim, Jonghyun and Kim, Joongkyu},
  booktitle={Proceedings of the IEEE/CVF conference on computer vision and pattern recognition},
  pages={8334--8343},
  year={2021}
}

@inproceedings{xu2023self,
  title={Self-calibrated cross attention network for few-shot segmentation},
  author={Xu, Qianxiong and Zhao, Wenting and Lin, Guosheng and Long, Cheng},
  booktitle={Proceedings of the IEEE/CVF international conference on computer vision},
  pages={655--665},
  year={2023}
}

@inproceedings{xu2024eliminating,
  title={Eliminating feature ambiguity for few-shot segmentation},
  author={Xu, Qianxiong and Lin, Guosheng and Loy, Chen Change and Long, Cheng and Li, Ziyue and Zhao, Rui},
  booktitle={European Conference on Computer Vision},
  pages={416--433},
  year={2024},
  organization={Springer}
}

@article{chang2024high,
  title={High-Performance Few-Shot Segmentation with Foundation Models: An Empirical Study},
  author={Chang, Shijie and Zhang, Lihe and Lu, Huchuan},
  journal={arXiv preprint arXiv:2409.06305},
  year={2024}
}

@article{xu2025unlocking,
  title={Unlocking the Power of SAM 2 for Few-Shot Segmentation},
  author={Xu, Qianxiong and Zhu, Lanyun and Liu, Xuanyi and Lin, Guosheng and Long, Cheng and Li, Ziyue and Zhao, Rui},
  journal={arXiv preprint arXiv:2505.14100},
  year={2025}
}

@inproceedings{kan2016multi,
  title={Multi-view deep network for cross-view classification},
  author={Kan, Meina and Shan, Shiguang and Chen, Xilin},
  booktitle={Proceedings of the IEEE conference on computer vision and pattern recognition},
  pages={4847--4855},
  year={2016}
}

@techreport{arthur2006k,
  title={k-means++: The advantages of careful seeding},
  author={Arthur, David and Vassilvitskii, Sergei},
  year={2006},
  institution={Stanford}
}

@inproceedings{demir2018deepglobe,
  title={Deepglobe 2018: A challenge to parse the earth through satellite images},
  author={Demir, Ilke and Koperski, Krzysztof and Lindenbaum, David and Pang, Guan and Huang, Jing and Basu, Saikat and Hughes, Forest and Tuia, Devis and Raskar, Ramesh},
  booktitle={Proceedings of the IEEE conference on computer vision and pattern recognition workshops},
  pages={172--181},
  year={2018}
}

@article{codella2019skin,
  title={Skin lesion analysis toward melanoma detection 2018: A challenge hosted by the international skin imaging collaboration (isic)},
  author={Codella, Noel and Rotemberg, Veronica and Tschandl, Philipp and Celebi, M Emre and Dusza, Stephen and Gutman, David and Helba, Brian and Kalloo, Aadi and Liopyris, Konstantinos and Marchetti, Michael and others},
  journal={arXiv preprint arXiv:1902.03368},
  year={2019}
}

@article{tschandl2018ham10000,
  title={The HAM10000 dataset, a large collection of multi-source dermatoscopic images of common pigmented skin lesions},
  author={Tschandl, Philipp and Rosendahl, Cliff and Kittler, Harald},
  journal={Scientific data},
  volume={5},
  number={1},
  pages={1--9},
  year={2018},
  publisher={Nature Publishing Group}
}

@article{candemir2013lung,
  title={Lung segmentation in chest radiographs using anatomical atlases with nonrigid registration},
  author={Candemir, Sema and Jaeger, Stefan and Palaniappan, Kannappan and Musco, Jonathan P and Singh, Rahul K and Xue, Zhiyun and Karargyris, Alexandros and Antani, Sameer and Thoma, George and McDonald, Clement J},
  journal={IEEE transactions on medical imaging},
  volume={33},
  number={2},
  pages={577--590},
  year={2013},
  publisher={IEEE}
}

@article{jaeger2013automatic,
  title={Automatic tuberculosis screening using chest radiographs},
  author={Jaeger, Stefan and Karargyris, Alexandros and Candemir, Sema and Folio, Les and Siegelman, Jenifer and Callaghan, Fiona and Xue, Zhiyun and Palaniappan, Kannappan and Singh, Rahul K and Antani, Sameer and others},
  journal={IEEE transactions on medical imaging},
  volume={33},
  number={2},
  pages={233--245},
  year={2013},
  publisher={IEEE}
}

@inproceedings{li2020fss,
  title={Fss-1000: A 1000-class dataset for few-shot segmentation},
  author={Li, Xiang and Wei, Tianhan and Chen, Yau Pun and Tai, Yu-Wing and Tang, Chi-Keung},
  booktitle={Proceedings of the IEEE/CVF conference on computer vision and pattern recognition},
  pages={2869--2878},
  year={2020}
}

@article{everingham2010pascal,
  title={The pascal visual object classes (voc) challenge},
  author={Everingham, Mark and Van Gool, Luc and Williams, Christopher KI and Winn, John and Zisserman, Andrew},
  journal={International journal of computer vision},
  volume={88},
  number={2},
  pages={303--338},
  year={2010},
  publisher={Springer}
}

@inproceedings{lin2014microsoft,
  title={Microsoft coco: Common objects in context},
  author={Lin, Tsung-Yi and Maire, Michael and Belongie, Serge and Hays, James and Perona, Pietro and Ramanan, Deva and Doll{\'a}r, Piotr and Zitnick, C Lawrence},
  booktitle={European conference on computer vision},
  pages={740--755},
  year={2014},
  organization={Springer}
}

@article{liu2023matcher,
  title={Matcher: Segment anything with one shot using all-purpose feature matching},
  author={Liu, Yang and Zhu, Muzhi and Li, Hengtao and Chen, Hao and Wang, Xinlong and Shen, Chunhua},
  journal={arXiv preprint arXiv:2305.13310},
  year={2023}
}

@article{zhang2023personalize,
  title={Personalize segment anything model with one shot},
  author={Zhang, Renrui and Jiang, Zhengkai and Guo, Ziyu and Yan, Shilin and Pan, Junting and Ma, Xianzheng and Dong, Hao and Gao, Peng and Li, Hongsheng},
  journal={arXiv preprint arXiv:2305.03048},
  year={2023}
}

@inproceedings{zhu2025maup,
  title={MAUP: Training-Free Multi-center Adaptive Uncertainty-Aware Prompting for Cross-Domain Few-Shot Medical Image Segmentation},
  author={Zhu, Yazhou and Zhang, Haofeng},
  booktitle={International Conference on Medical Image Computing and Computer-Assisted Intervention},
  pages={326--336},
  year={2025},
  organization={Springer}
}

@inproceedings{he2024apseg,
  title={Apseg: Auto-prompt network for cross-domain few-shot semantic segmentation},
  author={He, Weizhao and Zhang, Yang and Zhuo, Wei and Shen, Linlin and Yang, Jiaqi and Deng, Songhe and Sun, Liang},
  booktitle={Proceedings of the IEEE/CVF Conference on Computer Vision and Pattern Recognition},
  pages={23762--23772},
  year={2024}
}

@article{zhang2022feature,
  title={Feature-proxy transformer for few-shot segmentation},
  author={Zhang, Jian-Wei and Sun, Yifan and Yang, Yi and Chen, Wei},
  journal={Advances in neural information processing systems},
  volume={35},
  pages={6575--6588},
  year={2022}
}

@inproceedings{he2016deep,
  title={Deep residual learning for image recognition},
  author={He, Kaiming and Zhang, Xiangyu and Ren, Shaoqing and Sun, Jian},
  booktitle={Proceedings of the IEEE conference on computer vision and pattern recognition},
  pages={770--778},
  year={2016}
}

@article{dosovitskiy2020image,
  title={An image is worth 16x16 words: Transformers for image recognition at scale},
  author={Dosovitskiy, Alexey and Beyer, Lucas and Kolesnikov, Alexander and Weissenborn, Dirk and Zhai, Xiaohua and Unterthiner, Thomas and Dehghani, Mostafa and Minderer, Matthias and Heigold, Georg and Gelly, Sylvain and others},
  journal={arXiv preprint arXiv:2010.11929},
  year={2020}
}

@inproceedings{liu2025synpo,
  title={SynPo: Boosting Training-Free Few-Shot Medical Segmentation via High-Quality Negative Prompts},
  author={Liu, Yufei and Xiao, Haoke and Chai, Jiaxing and Zhang, Yongcun and Wang, Rong and Meng, Zijie and Luo, Zhiming},
  booktitle={International Conference on Medical Image Computing and Computer-Assisted Intervention},
  pages={594--603},
  year={2025},
  organization={Springer}
}

@inproceedings{fan2025adapting,
  title={Adapting In-Domain Few-Shot Segmentation to New Domains without Source Domain Retraining},
  author={Fan, Qi and Liu, Kaiqi and Liu, Nian and Cholakkal, Hisham and Anwer, Rao Muhammad and Li, Wenbin and Gao, Yang},
  booktitle={Proceedings of the IEEE/CVF International Conference on Computer Vision},
  pages={21429--21439},
  year={2025}
}

@article{ravi2024sam,
  title={Sam 2: Segment anything in images and videos},
  author={Ravi, Nikhila and Gabeur, Valentin and Hu, Yuan-Ting and Hu, Ronghang and Ryali, Chaitanya and Ma, Tengyu and Khedr, Haitham and R{\"a}dle, Roman and Rolland, Chloe and Gustafson, Laura and others},
  journal={arXiv preprint arXiv:2408.00714},
  year={2024}
}

@inproceedings{shaban2017one,
  title={One-Shot Learning for Semantic Segmentation},
  author={Shaban, Amirreza and Bansal, Shray and Liu, Zhen and Essa, Irfan and Boots, Byron},
  booktitle={Procedings of the British Machine Vision Conference 2017},
  year={2017},
  organization={British Machine Vision Association}
}

@inproceedings{sun2026bridging,
  title={Bridging Granularity Gaps: Hierarchical Semantic Learning for Cross-domain Few-shot Segmentation},
  author={Sun, Sujun and Gu, Haowen and Xie, Cheng and Ren, Yanxu and Ren, Mingwu and Zhang, Haofeng},
  booktitle={Proceedings of the AAAI Conference on Artificial Intelligence},
  volume={40},
  number={11},
  pages={9215--9223},
  year={2026}
}
\end{document}


\title{Training-free Cross-domain Few-shot Segmentation via Robust Semantic Representation and Matching} 

\titlerunning{Training-free CD-FSS via Robust Semantic Representation and Matching}

\author{Sujun Sun\inst{1,2}\orcidlink{0009-0002-0647-1148} \and
Mingwu Ren\inst{1,2}\orcidlink{0000-0001-5576-3281} \and
Haofeng Zhang\inst{1,2}\textsuperscript{(\Letter)}\orcidlink{0000-0002-4039-7618}}

\authorrunning{S. Sun et al.}

\institute{School of Computer Science and Engineering, Nanjing University of Science and Technology, China \and
State Key Laboratory of Intelligent Manufacturing of Advanced Construction Machinery, China\\
\email{\{egg, renmingwu, zhanghf\}@njust.edu.cn}}

\maketitlesupplementary

\section{Fisher Discriminant Ratio}
\label{sec:fdr}
Inspired by \cite{kan2016multi}, we use the Fisher discriminant ratio (FDR) to measure the semantic discriminability of features. The FDR is the trace ratio between the between-class and within-class scatter matrices, reflecting the inter-class discriminability of features. Specifically, given two sample sets $\mathcal{X}$ and $\mathcal{Y}$, the FDR $f$ is computed as follows:
\begin{equation}
    f=\frac{\operatorname{Tr}(\mathbf{S}_B)}{\operatorname{Tr}(\mathbf{S}_W)},
\end{equation}
where $\operatorname{Tr}(\cdot)$ denotes the trace of a matrix, $\mathbf{S}_B$ and $\mathbf{S}_W$ represent the between-class and within-class scatter matrices, respectively. The trace operation provides a scalar measure of the total scatter, making this formulation computationally efficient and numerically stable. The between-class scatter $\mathbf{S}_B$ captures the separation between distribution centers:
\begin{equation}
    \mathbf{S}_B=(\boldsymbol{\mu}_x-\boldsymbol{\mu}_y)(\boldsymbol{\mu}_x-\boldsymbol{\mu}_y)^{T},
\end{equation}
where $\boldsymbol{\mu}_x$ and $\boldsymbol{\mu}_y$ are the mean vectors of distributions $\mathcal{X}$ and $\mathcal{Y}$, respectively. The within-class scatter $\mathbf{S}_W$ measures the internal variance:
\begin{equation}
    \mathbf{S}_W=\boldsymbol{\Sigma}_x+\boldsymbol{\Sigma}_y,
\end{equation}
where $\boldsymbol{\Sigma}_x$ and $\boldsymbol{\Sigma}_y$ are the covariance matrices of distributions $\mathcal{X}$ and $\mathcal{Y}$, respectively.

Given a feature map, we treat all foreground and background features as distributions $\mathcal{X}$ and $\mathcal{Y}$, and use FDR to quantify the separation between foreground and background, thereby reflecting its semantic discriminability. For cross-image (support and query) features, we compute the FDR between support background and query foreground, as well as between support foreground and query foreground, and use the difference between the two to reflect the cross-image semantic discriminability.

\begin{figure}[t]
\centering
    \includegraphics[width=\linewidth]{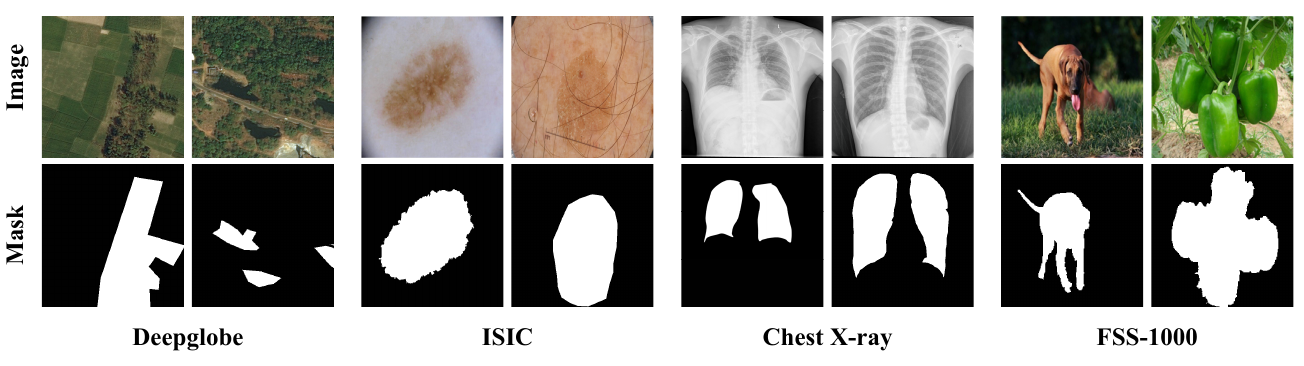}
    \caption{Images and their corresponding ground truth masks sampled from four target domain datasets.}
    \label{fig:samples}
\end{figure}

\section{Dataset Details}
\label{sec:dataset}
Following the setup of PATNet ~\cite{lei2022cross}, we evaluate our method on four target domain datasets: Deepglobe ~\cite{demir2018deepglobe}, ISIC ~\cite{codella2019skin,tschandl2018ham10000}, Chest X-ray ~\cite{candemir2013lung,jaeger2013automatic}, and FSS-1000 ~\cite{li2020fss}. These datasets cover diverse remote sensing, medical, and natural images, providing rich cross-domain diversity and posing challenges to model generalization. \cref{fig:samples} shows sampled images from each dataset.

\textbf{Deepglobe} consists of high-resolution satellite remote sensing images with pixel-level annotations covering seven categories: urban, agriculture, rangeland, forest, water, barren, and unknown. Following PATNet ~\cite{lei2022cross}, we crop the original images of resolution $2448 \times 2448$ into 6 sub-images and filter out single-class images and the `unknown' class.

\textbf{ISIC} is a medical image dataset, containing three types of skin lesion images with corresponding pixel-level annotations. Following common practice, we resize the images from the original resolution of $1022 \times 767$ to $512 \times 512$.

\textbf{Chest X-ray} is an X-ray imaging dataset for tuberculosis diagnosis. It contains 566 medical images with an original resolution of $4020 \times 4892$, collected from 58 confirmed tuberculosis patients and 80 healthy individuals. All images are uniformly downsampled to $1024 \times 1024$ for testing.

\textbf{FSS-1000} is a dataset specifically designed for few-shot segmentation. It contains 1000 target categories from natural scenes and daily objects, each with 10 samples annotated at the pixel level. Following the official data split protocol, we evaluate on the test set, which includes 2400 images from 240 categories.

\begin{figure}[t]
\centering
    \includegraphics[width=1.0\linewidth]{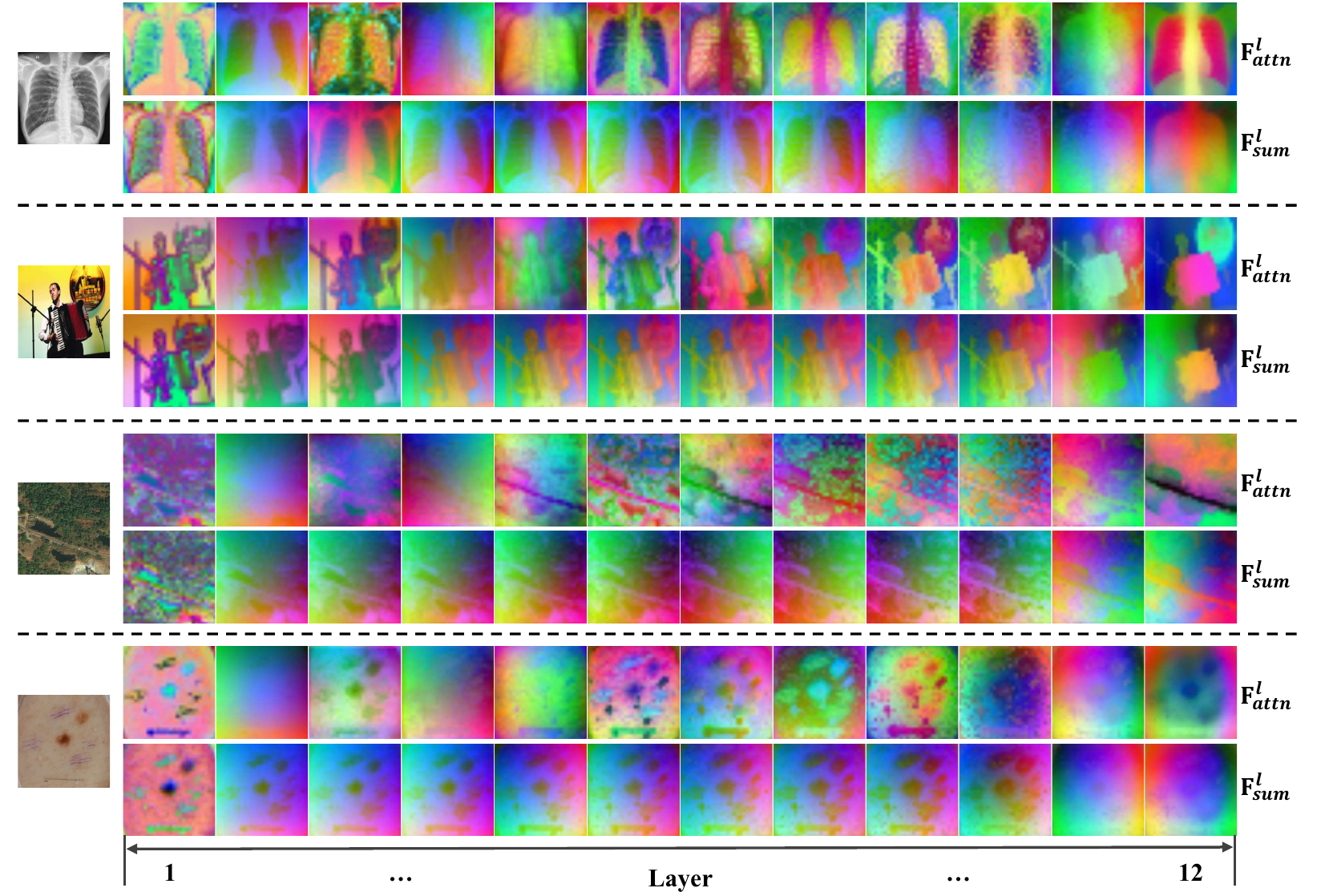}
    \caption{Visualization of the raw attention outputs and fused outputs after residual connections in each DINOv3 block. Layers focusing on semantic patterns and local consistency appear alternately, and the position-aware features that emphasize local consistency cause the fused output $\mathbf{F}^l_{sum}$ to also overly focus on local consistency, which persists to the final output.}
    \label{fig:layers_pca}
\end{figure}

\begin{figure}[t!]
\centering
    \includegraphics[width=1.0\linewidth]{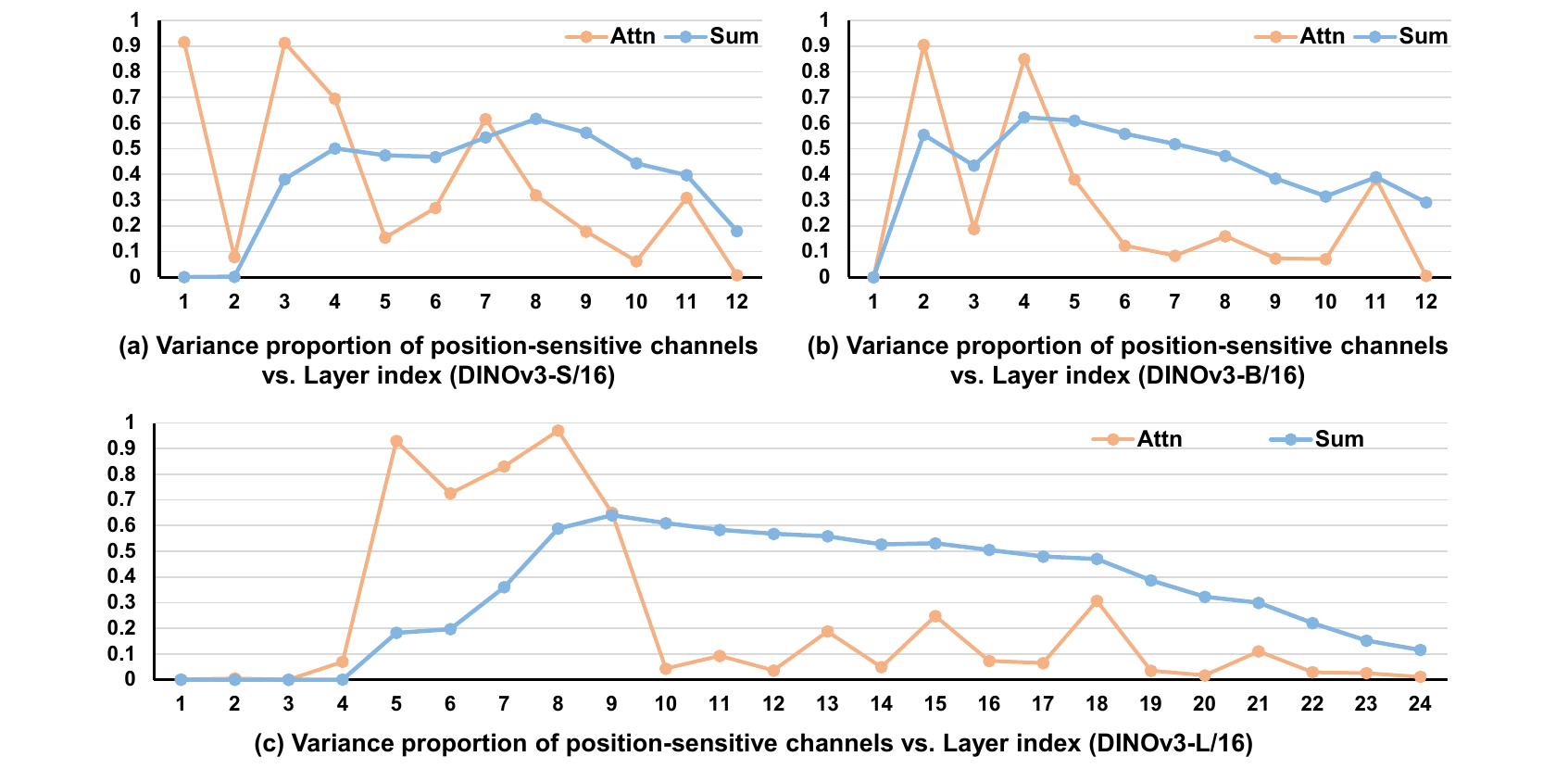}
    \caption{Variance proportion of position-sensitive channels in each layer across different DINOv3 architectures.}
    \label{fig:layers_analysis}
\end{figure}

\section{Analysis of Layer-wise Features in DINOv3}
\label{sec:feat_analysis}
We visualize the layer-wise features of DINOv3-B/16 \cite{simeoni2025dinov3}, including the raw outputs $\{\mathbf{F}^l_{attn}\}_{l=1}^L$ of attention modules and the fused outputs $\{\mathbf{F}^l_{sum}\}_{l=1}^L$ after residual connections in each block, as shown in \cref{fig:layers_pca}. During feature extraction, some layers focusing on semantic information may introduce noise, while subsequent layers emphasizing local consistency help smooth the learned semantics. From shallow to deep layers, the network alternately attends to semantic patterns and local consistency, extracting semantic-aware and position-aware features, respectively. However, after position-aware features are fused into $\mathbf{F}^l_{sum}$, $\mathbf{F}^l_{sum}$ becomes overly focused on local consistency, and this effect persists to the final output $\mathbf{F}_{last} = \mathbf{F}^L_{sum}$. We further calculate the proportion of variance from position-sensitive channels to the total variance in each layer, as shown in \cref{fig:layers_analysis}(b). Position-aware features exhibit a higher variance proportion, which increases the corresponding variance proportion in $\mathbf{F}^l_{sum}$ and persists to the final output.

As shown in \cref{fig:layers_analysis}, this issue is observed in different DINOv3 architectures. Nevertheless, as the network depth increases, the influence of position-aware features on $\mathbf{F}^l_{sum}$ gradually decreases. Moreover, at the same network depth, the final output of DINOv3-S is less affected by position-sensitive channels compared to DINOv3-B. We attribute this to the fewer feature channels in DINOv3-S, which introduces less semantic noise and a weaker ability of layers that focus on local consistency. In \cref{tab:arch}, the baseline achieves higher performance with DINOv3-S than with DINOv3-B, which is consistent with the observation that DINOv3-S is less affected by position-sensitive channels. In contrast, our method performs worse with DINOv3-S, validating its weaker learning ability.

\begin{table}[t]
    \caption{Comparison with existing training-free FSS methods on PASCAL-$5^i$.}
    \label{tab:performance_std}
    \centering
    \setlength{\tabcolsep}{2.4mm}
    \scalebox{.85}{
        \begin{tabular}{c|c|ccccc}
            \toprule
            Method & Mark & $5^0$ & $5^1$ & $5^2$ & $5^3$ & Average\\
            \midrule
            PerSAM \cite{zhang2023personalize} & ICLR-24 & - & - & - & - & 43.1 \\
            Matcher \cite{liu2023matcher} & ICLR-24 & 67.7 & 70.7 & 66.9 & 67.0 & 68.1 \\
            GF-SAM \cite{zhang2024bridge} & NIPS-24 & 71.1 & 75.7 & 69.2 & 73.3 & 72.1 \\
            \textbf{Ours} & Ours & 71.4 & 74.6 & 74.7 & 72.0 & \textbf{73.2} \\
            \bottomrule
        \end{tabular}}
\end{table}

\section{Comparison on the Standard FSS Dataset}
\label{sec:std_fss}
We compare our method with training-free FSS methods \cite{zhang2023personalize, liu2023matcher, zhang2024bridge} on the standard FSS dataset PASCAL-$5^i$ \cite{shaban2017one}, with 1-shot results reported in \cref{tab:performance_std}. Although designed for generalization to diverse domains, our method still outperforms existing methods on PASCAL-$5^i$, which contains complex natural images.

\section{More Ablation Studies}
\label{sec:ablation}
We provide more method comparison and hyperparameter analysis through extensive experiments. Unless otherwise specified, all experiments are conducted under 1-shot setting.

\begin{table}[ht]
    \caption{Ablation studies on more pretrained models.}
    \label{tab:pretrained}
    \centering
    \setlength{\tabcolsep}{2.4mm}
    \scalebox{.85}{
        \begin{tabular}{l|ccccc}
            \toprule
             & ViT-B/16 & DINOv2-B/14 & DINOv3-B/16 & SAM-B/16 & SAM2-B/16 \\
            \midrule
            Baseline & 54.94 & 55.02 & \textbf{58.74} & 53.94 & 56.05 \\
            Ours & 58.92 & 61.38 & \textbf{68.39} & 58.60 & 61.86 \\
            \bottomrule
        \end{tabular}}
    \vspace{-1ex}
\end{table}

\noindent
\textbf{Ablation on more pretrained models.}
We use ViT-B/16 \cite{dosovitskiy2020image} (with ImageNet \cite{russakovsky2015imagenet} pretrained weights), DINOv2-B/14 \cite{oquab2023dinov2}, DINOv3-B/16 \cite{simeoni2025dinov3}, SAM-B/16 \cite{kirillov2023segment}, and SAM2-B/16 \cite{ravi2024sam} as backbones to validate the effectiveness of our method across different pretrained models, as shown in \cref{tab:pretrained}. It could be observed: 1) Our method achieves significant performance improvements across all backbones. 2) DINOv3 exhibits the best performance on the CD-FSS benchmark. 3) SAM performs even worse than the ViT-B in CD-FSS, indicating its limited cross-image semantic matching capability, while SAM2 alleviates this limitation by incorporating semantic consistency across video frames.

\begin{table}[htb]
    \caption{Ablation studies on different model architectures.}
    \label{tab:arch}
    \centering
    \setlength{\tabcolsep}{2.4mm}
    \scalebox{.85}{
        \begin{tabular}{c|c|ccccc}
            \toprule
            Backbone & Method & Deepglobe & ISIC & Chest X-ray & FSS-1000 & Average\\
            \midrule
            \multirow{2}*{DINOv3-S} & Baseline & 41.87 & 51.11 & 77.31 & 74.43 & 61.18 \\
            & Ours & 46.33 & 54.23 & 84.13 & 78.85 & \textbf{65.89} \\
            \midrule
            \multirow{2}*{DINOv3-B} & Baseline & 42.32 & 51.15 & 67.18 & 74.32 & 58.74 \\
            & Ours & 49.71 & 55.73 & 85.44 & 82.67 & \textbf{68.39} \\
            \midrule
            \multirow{2}*{DINOv3-L} & Baseline & 45.31 & 54.80 & 76.93 & 72.55 & 62.40 \\
            & Ours & 48.73 & 57.83 & 87.43 & 80.88 & \textbf{68.72} \\
            \bottomrule
        \end{tabular}}
    \vspace{-1ex}
\end{table}

\noindent
\textbf{Ablation on different architectures.}
To evaluate the effectiveness of our method on different architectures, we replace the backbone with DINOv3-S/16, DINOv3-B/16, and DINOv3-L/16, respectively. As shown in \cref{tab:arch}, our method achieves significant performance improvements on all architectures.

\begin{table}[htb]
    \caption{Comparison with other training-free matching strategies.}
    \label{tab:tr_match}
    \centering
    \setlength{\tabcolsep}{2.4mm}
    \scalebox{.85}{
        \begin{tabular}{c|ccccc}
            \toprule
            Method & Deepglobe & ISIC & Chest X-ray & FSS-1000 & Average\\
            \midrule
            Baseline & 42.32 & 51.15 & 67.18 & 74.32 & 58.74 \\
            DINOv3+SSP \cite{fan2022self} & 40.41 & 54.61 & 59.47 & 79.24 & 58.43 \\
            DINOv3+IFA \cite{nie2024cross} & 40.97 & 50.22 & 49.19 & 75.71 & 54.02 \\
            Ours & 49.71 & 55.73 & 85.44 & 82.67 & \textbf{68.39} \\
            \bottomrule
        \end{tabular}}
    \vspace{-1ex}
\end{table}

\noindent
\textbf{Comparison with training-free matching strategies.}
Some FSS and CD-FSS methods train the backbone and design training-free support-query matching strategies to enhance model performance. We replace the backbone in these methods with a frozen DINOv3-B/16 and employ their training-free matching strategies to obtain query predictions. Specifically, we compare with the training-free matching strategies of SSP \cite{fan2022self} and IFA \cite{nie2024cross} to further validate the effectiveness of our training-free modules, as shown in \cref{tab:tr_match}. These methods typically rely on fixed thresholds and fail to generalize across different domains or backbones, resulting in lower performance than the baseline. In contrast, our method significantly outperforms these methods.

\begin{table}[htb]
    \caption{Ablation studies for feature fusion strategies.}
    \label{tab:fusion_strategy}
    \centering
    \setlength{\tabcolsep}{2.4mm}
    \scalebox{.85}{
        \begin{tabular}{c|ccccc}
            \toprule
            Strategy & Deepglobe & ISIC & Chest X-ray & FSS-1000 & Average\\
            \midrule
            All layers & 47.30 & 53.23 & 71.19 & 74.87 & 61.65 \\
            All layers w/o $\mathcal{L}_{pos}$ & 48.26 & 53.54 & 71.22 & 74.95 & 61.99 \\
            $\mathcal{L}_{sem}$ & 48.78 & 53.69 & 72.03 & 75.60 & \textbf{62.53} \\
            Optimal layer & 48.93 & 53.48 & 70.93 & 75.31 & 62.16 \\
            \bottomrule
        \end{tabular}}
    \vspace{-1ex}
\end{table}

\noindent
\textbf{Effects of feature fusion strategies.}
As shown in \cref{tab:fusion_strategy}, we explore the effects of features from different layers by comparing various feature fusion strategies. First, we fuse the raw outputs $\{\mathbf{F}^l_{attn}\}_{l=1}^L$ from all layers. Removing the position-aware layers in $\mathcal{L}_{pos}$ improves performance, indicating that even low-weight position-aware features can reduce semantic discriminability. Further improvement is observed when fusing representative semantic-aware layers $\mathcal{L}_{sem}$, outperforming the use of only the optimal single layer. This indicates that semantic-aware features from different stages provide additional foreground-background discrimination to the optimal feature.

\begin{table}[htb]
    \caption{Ablation studies for the effects of weighting strategies.}
    \label{tab:weighting_strategy}
    \centering
    \setlength{\tabcolsep}{2.4mm}
    \scalebox{.85}{
        \begin{tabular}{c|ccccc}
            \toprule
            Strategy & Deepglobe & ISIC & Chest X-ray & FSS-1000 & Average\\
            \midrule
            Only $r_{attn}$ & 48.67 & 53.48 & 71.04 & 75.41 & 62.15 \\
            Only $f_{attn}$ & 47.28 & 52.50 & 69.06 & 69.47 & 59.58 \\
            Combined weights & 48.78 & 53.69 & 72.03 & 75.60 & \textbf{62.53} \\
            \bottomrule
        \end{tabular}}
    \vspace{-1ex}
\end{table}

\noindent
\textbf{Effects of weighting strategies.}
We conduct additional ablation studies to explore the effects of different weighting strategies in SAFR, with results shown in \cref{tab:weighting_strategy}. Specifically, we compute the feature fusion weights using only the variance proportion $r_{attn}$ of position-sensitive channels or only the FDR $f_{attn}$. The results show that using $r_{attn}$ achieves better performance. We attribute this to features with lower $r_{attn}$ focusing more on semantic patterns and possessing stronger intra-image and inter-image semantic discriminability, while the FDR $f_{attn}$ only reflects the semantic discriminability within a single image. However, $f_{attn}$ can serve as complementary information, and combining both leads to improved performance.

\begin{table}[htb]
    \vspace{-1ex}
    \caption{Ablation studies for the effects of different $\gamma$.}
    \label{tab:gamma}
    \centering
    \setlength{\tabcolsep}{2.4mm}
    \scalebox{.85}{
        \begin{tabular}{c|cccc}
            \toprule
            $\gamma$ & 100 & 200 & 300 & 400\\
            \midrule
            mean-IoU & 62.51 & \textbf{62.53} & 62.51 & 62.49 \\
            \bottomrule
        \end{tabular}}
    \vspace{-1ex}
\end{table}

\noindent
\textbf{Effects of different $\gamma$.}
We set different values for $\gamma$ in SAFR to control the number $N_c$ of selected position-sensitive channels and explore its impact on the results. As shown in \cref{tab:gamma}, varying $\gamma$ has little effect on the results, and the model achieves the best performance when $\gamma$ is set to 200.

\begin{figure}[t]
\centering
    \includegraphics[width=\linewidth]{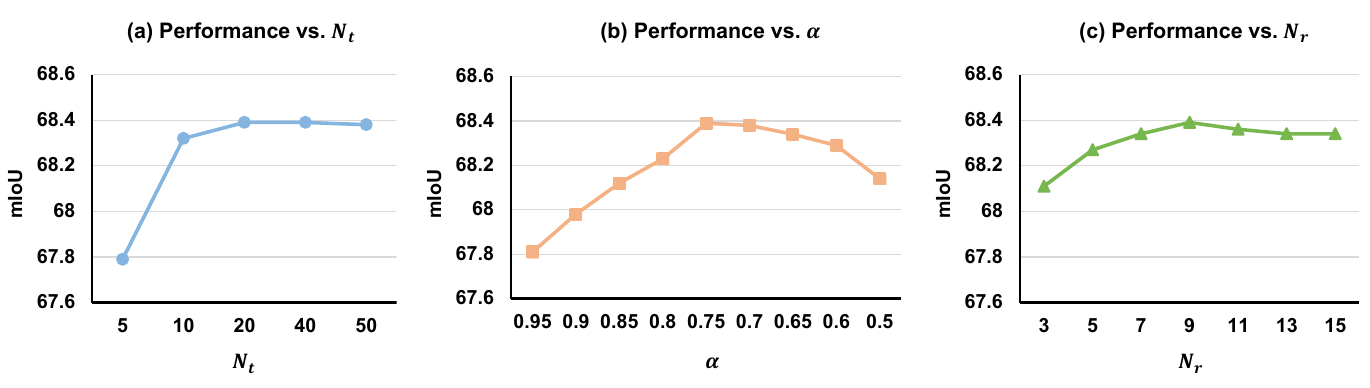}
    \caption{Parameter studies. (a) The mIoU curves over different numbers of sampled thresholds in ASE. (b) The mIoU curves over different coefficients $\alpha$ in ASE. (c) The mIoU curves over different numbers of foreground regional prototypes in HPM.}
    \label{fig:parameter}
\end{figure}

\begin{table}[htb]
    \caption{Generalization studies of $N_t$ setting across different backbones.}
    \label{tab:N_t}
    \centering
    \setlength{\tabcolsep}{2.4mm}
    \scalebox{.85}{
        \begin{tabular}{c|ccccc}
            \toprule
            $N_t$ & 5 & 10 & 20 & 40 & 50 \\
            \midrule
            DINOv3 & 67.79 & 68.32 & \textbf{68.39} & 68.39 & 68.38 \\
            DINOv2 & 61.25 & 61.38 & \textbf{61.38} & 61.38 & 61.36 \\
            SAM2 & 61.60 & 61.82 & 61.86 & \textbf{61.87} & 61.86 \\
            \bottomrule
        \end{tabular}}
    \vspace{-4ex}
\end{table}

\begin{table}[htb]
    \vspace{-2ex}
    \caption{Generalization studies of $\alpha$ setting across different backbones.}
    \label{tab:alpha}
    \centering
    \setlength{\tabcolsep}{2.4mm}
    \scalebox{.85}{
        \begin{tabular}{c|ccccccccc}
            \toprule
            $\alpha$ & 0.95 & 0.9 & 0.85 & 0.8 & 0.75 & 0.7 & 0.65 & 0.6 & 0.5 \\
            \midrule
            DINOv3 & 67.81 & 67.98 & 68.12 & 68.23 & \textbf{68.39} & 68.38 & 69.34 & 68.29 & 68.14 \\
            DINOv2 & 61.27 & 61.32 & 61.33 & 61.36 & 61.38 & \textbf{61.40} & 61.35 & 61.28 & 61.12 \\
            SAM2 & 61.72 & 61.74 & 61.75 & 61.79 & \textbf{61.86} & 61.79 & 61.77 & 61.75 & 61.64 \\
            \bottomrule
        \end{tabular}}
\end{table}

\noindent
\textbf{Hyperparameter analysis in ASE.}
We conduct ablation studies on hyperparameters in ASE, including the number $N_t$ of sampled thresholds and the robust threshold coefficient $\alpha$. As shown in \cref{fig:parameter}(a), when $N_t \geq 10$ (\textit{i.e.}, the threshold interval is less than or equal to 0.1), a precise estimation of the robust threshold can be obtained. We finally set $N_t$ to 20. As shown in \cref{fig:parameter}(b), the best performance is achieved when $\alpha$ is set to 0.75. We further analyze the generalization of the $N_t$ and $\alpha$ settings. As shown in \cref{tab:N_t} and \cref{tab:alpha}, the impact of $N_t$ and $\alpha$ on performance exhibits consistent trends across different backbones.

\noindent
\textbf{Effects of different $N_r$.}
As shown in \cref{fig:parameter}(c), we evaluate model performance with different numbers of foreground regional prototypes. Fewer regional prototypes improve performance on datasets that favor global prototypes, while more regional prototypes enhance performance on datasets that favor pixel-level prototypes. The model achieves the best balance when $N_r$ is set to 9. We further analyze the generalization of the $N_r$ setting, and \cref{tab:N_r} shows that the impact of $N_r$ on performance exhibits consistent trends across different backbones.

\begin{table}[htb]
    \caption{Generalization studies of $N_r$ setting across different backbones.}
    \label{tab:N_r}
    \centering
    \setlength{\tabcolsep}{2.4mm}
    \scalebox{.85}{
        \begin{tabular}{c|ccccccccc}
            \toprule
            $N_r$ & 3 & 5 & 7 & 9 & 11 & 13 & 15 \\
            \midrule
            DINOv3 & 68.11 & 68.27 & 68.34 & \textbf{68.39} & 68.36 & 68.34 & 68.34 \\
            DINOv2 & 61.28 & 61.30 & 61.38 & \textbf{61.38} & 61.37 & 61.31 & 61.32 \\
            SAM2 & 61.72 & 61.81 & 61.80 & \textbf{61.86} & 61.82 & 61.80 & 61.77 \\
            \bottomrule
        \end{tabular}}
\end{table}

\section{Limitations and Future Work}
\label{sec:limitation}
When instances in query and support differ significantly, our method still faces the challenge of insufficient support information. Effectively leveraging textual information (\textit{e.g.}, category descriptions) to enrich support in cross-domain scenarios is a direction for future research.

%
%
\bibliographystyle{splncs04}
\bibliography{main}